\documentclass[12pt,journal,compsoc]{IEEEtran}
\usepackage{lineno,hyperref}
\modulolinenumbers[5]

\ifCLASSOPTIONcompsoc
   \usepackage[nocompress]{cite}
\else

   \usepackage{cite}
\fi
\usepackage[pdftex]{graphicx}
\usepackage[cmex10]{amsmath}
\usepackage{amsthm}
\usepackage{bm}
\usepackage{algorithm}
\usepackage{algpseudocode}
\usepackage{lipsum}
\usepackage{varwidth}

\newtheorem{definition}{Definition}
\usepackage{array}
\usepackage[font=footnotesize]{subfig}
\usepackage{fixltx2e}
\usepackage{color}
\usepackage{amsfonts}

\newtheorem{thm}{Theorem}

\begin{document}

\title{Causality on Cross-Sectional Data:\\ Stable Specification Search in Constrained Structural Equation Modeling}
\author{Ridho~Rahmadi, Perry~Groot, Marianne~Heins, \\Hans~Knoop, Tom~Heskes, The OPTIMISTIC~consortium$^*$
\IEEEcompsocitemizethanks{\IEEEcompsocthanksitem R. Rahmadi is with the Department
of Informatics, Universitas Islam Indonesia and Institute for Computing and Information Sciences, Radboud University Nijmegen, the Netherlands. E-mail: r.rahmadi@cs.ru.nl 
\IEEEcompsocthanksitem P. Groot and T. Heskes are with Institute for Computing and Information Sciences, Radboud University Nijmegen, the Netherlands.
\IEEEcompsocthanksitem M. Heins and H. Knoop are with Expert Centre for Chronic Fatigue, Radboud University Medical Centre, the Netherlands.
\IEEEcompsocthanksitem $^*$The members of OPTIMISTIC consortium are described in~\cite{van2015cognitive}.
}}

\IEEEtitleabstractindextext{
\begin{abstract}
Causal modeling has long been an attractive topic for many researchers and in recent decades
there has seen a surge in theoretical development and discovery algorithms. Generally discovery algorithms can be divided into two approaches: constraint-based and score-based. The constraint-based approach is able to detect common causes of the observed
variables but the use of independence tests makes it less reliable. The score-based approach produces a result that is easier to interpret as
it also measures the reliability of the inferred causal relationships, but it is unable to detect common confounders of the observed variables.
A drawback of both score-based and constrained-based approaches is the inherent instability in structure estimation.
With finite samples small changes in the data can lead to completely different optimal structures. The present work introduces a new hypothesis-free score-based causal discovery algorithm, called stable specification search, that is robust for finite samples based on recent advances in stability selection using subsampling and selection algorithms. Structure search is performed
over Structural Equation Models. Our approach uses exploratory search but allows incorporation of prior background knowledge.
We validated our approach on one simulated data set, which we compare to the known ground truth, and two real-world data sets for Chronic Fatigue Syndrome and Attention Deficit Hyperactivity Disorder, which we compare to earlier medical studies. The results on the simulated data set show significant improvement over alternative approaches and the results on the real-word data sets show consistency with the hypothesis driven models constructed by medical experts.
\end{abstract}

}

\maketitle
\IEEEdisplaynontitleabstractindextext
\IEEEpeerreviewmaketitle

\section{Introduction}
\IEEEPARstart{C}{ausal} modeling has been an attractive topic for many researchers for decades. Especially since the 1990s there has been an enormous increase in theoretical development, partly because of advances in graphical modeling \cite{pearl2000causality}. This has led to a variety of causal discovery algorithms in the literature. In general, causal discovery algorithms can be divided into two approaches: constraint-based and score-based. Constraint-based approaches work with conditional independence tests. First, they construct a skeleton graph starting with the complete graph and excluding edges between variables that are conditionally independent. Second, edges are oriented to arrive at a causal graph. Examples of constraint-based approaches are the IC algorithm \cite{pearl1991theory}, PC-FCI \cite{spirtes2000causation}, and TC \cite{pellet2008using}. Constraint-based approaches do not have to rely on the
causal sufficiency assumption, and then can detect common causes of the observed variables \cite{spirtes2000causation}.
A disadvantage of this approach is the use of independence tests on a large number of conditioning variables, making it less reliable \cite{spirtes2010introduction}.
Score-based approaches assign scores to particular graph structures based on the data fit and the complexity of the graph. Different scoring metrics that are often used are the Bayesian score \cite{dawid1984present} and the BIC score \cite{schwarz1978estimating}. An example of a score-based method is \emph{greedy equivalence search} (GES) \cite{chickering2002optimal}. The goal of the score-based approach is to find the graph structure with the highest score. An advantage of this approach is that it measures the reliability of the inferred causal relationships, which makes the result easy to interpret \cite{heckerman1999bayesian}.
Score-based approaches typically do make the causal sufficiency assumption, and then cannot detect
common confounders of the observed variables. Moreover, the involved optimization problem is usually NP-hard, so that different search heuristics are often used.
The approach advocated in this paper is an example of a score-based approach.

Furthermore, in causal modeling based on observational data, the causal models are undetermined unless a preference for parsimonious models over more complex models is made \cite{spirtes2010introduction}. In score-based approaches, such simplicity assumptions are typically implemented by adding a penalty for model complexity \cite{chickering2002optimal}. Constraint-based approaches often make the implicit assumption of so-called \emph{causal faithfulness} \cite{spirtes2010introduction}, which states that there are no conditional independencies that hold in the density over a set of variables $V$, except those that are entailed by the causal structure. However, in practice faithfulness can be violated and better constrained-based approaches have been developed to handle this, such as CPC \cite{ramsey2012adjacency} and ACPC \cite{lemeire2012conservative}.

A drawback of both score-based and constrained-based approaches, however, is the inherent instability in structure estimation. With finite samples small changes in the data can lead to completely different optimal structures. Outcomes of borderline independence tests can be incorrect and can lead to multiple errors when propagated by the discovery algorithm \cite{spirtes2010introduction}.

The present work introduces a new score-based causal discovery algorithm, called \emph{stable specification search}, that is robust for finite samples based on advances in stability selection using subsampling and selection algorithms. Structure search is performed over Structural Equation Models (SEM), which is the most widely used language for causal discovery in various scientific disciplines. The method uses exploratory search, but allows incorporation of prior background knowledge.
In order to show that our method can handle various kinds of data (continuous, discrete, and a combination of both) we evaluated our method on simulated and real-world data. The simulated data is used to compare our method with some advanced constrained-based approaches (PC-stable \cite{colombo2014order}, CPC) and a score-based approach (GES).
Specifically, we compare the robustness of each method in computing causal structure. The real-world data sets, about
Chronic Fatigue Syndrome and Attention Deficit Hyperactivity Disorder, are used to compare our results with some previous studies. The results show that our exploratory, hypothesis-free approach gives significant improvement over alternative approaches, and is able to obtain structure estimates that are consistent with the hypothesis driven models constructed by medical experts based on medical data and years of experience.

The rest of this paper is structured as follows. Section~\ref{sec:background} describes all the background material obtained from the existing literature. Section~\ref{sec:proposedMethod} describes our robust score-based approach for causal discovery. Section~\ref{sec:experimental} presents experimental result on one simulated and two real-world data sets. Section~\ref{sec:conclusion} gives conclusions and \mbox{suggestions} for future work.

\section{Background}
\label{sec:background}
\subsection{Directed Acyclic Graph}
\label{sec:DAG}
\begin{figure}[!b]
\centering
\includegraphics[width=0.35\textwidth]{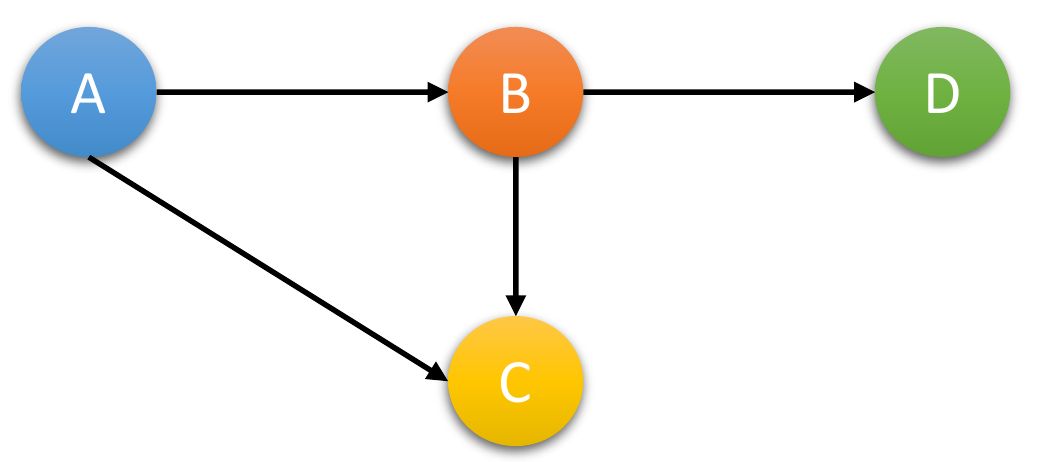}
\caption{A DAG of four variables.}
\label{DAG}
\end{figure}

We first describe some graphical notation and terminology used in the remaining sections.
A graph is a pair $(V,E)$ with $V$ a set of nodes and $E$ a set of edges. A \emph{directed graph} has all edges in $E$ directed (arc); a single arrowhead on every edge, e.g., \mbox{$A\to B$}. Directed cycles represent feedback or reciprocal relationships, e.g., $A\to B\to A$. A graph with no directed cycles is called \emph{acyclic}. A graph which is both directed and acyclic is called a \emph{Directed Acyclic Graph} (DAG)  \cite{pearl2000causality}. Figure \ref{DAG} depicts a DAG of four variables.
The \emph{skeleton} of a DAG is the undirected graph that results from removing the directionality of every edge. A \emph{v-structure} in a DAG $G$ is an ordered triple $(x,y,z)$ such that $G$ contains the directed edges $x\to y$, $z\to y$, and $x$ and $z$ are not adjacent in $G$ \cite{chickering2002learning}.

\subsection{Causal Modeling in SEM}
\label{subsec:SEM}
In this study, we focus on causal models with no reciprocal or feedback relationships, and no latent variables. Generally, there are two common ways of representing a model in SEM: by stating all relations in the set as equations, which is called a \emph{causal model}, or by drawing them as a \emph{causal diagram} (graph). The general form of the equations is
\begin{equation}
\label{SEM_eq2}
    x_i=f_i(\mathrm{pa}_i,\varepsilon_i),\quad i=1,\ldots,n.
\end{equation}
where $\mathrm{pa}_i$ denotes the \emph{parents} which represent the set of variables considered to be direct causes of $X_i$ and $\varepsilon_i$ represents errors on account of omitted factors that are assumed to be mutually independent \cite{pearl2000causality}.

\subsection{Specification Search in SEM}
Typically, a SEM is used as follows: 1) set a hypothesis as the prior model, 2) fit the model to the data, 3) evaluate the model, and 4) modify the model to improve the parsimony and score \cite{maccallum1986specification}. The last step is called \emph{specification search} \cite{leamer1978specification,long1983covariance}. This typical model refinement approach is hypothesis-driven. It works by adding or deleting some arcs between variables from the initial model. Typically only a few models are evaluated, making it difficult to derive causal relationships.

An alternative approach is exploratory search in which no prior hypothesis is specified. Typical approaches in the literature for addressing the exponential search space include tabu search \cite{marcoulides1998model}, genetic algorithms \cite{marcoulides2001specification,murohashi2007model}, ant colony optimization \cite{marcoulides2003model}, and others \cite{herting1985respecification,spirtes1990simulation,saris1987detection}.

\subsection{Multi-objective Optimization}
\label{sec:multi-objective}
Following the principle of Occam's razor, we should prefer models that are simple and fit the data well. These two objectives, however, are often conflicting as a well-fit model is likely to be a complex model. In this paper, we propose to make use of multi-objective optimization to explicitly optimize both objectives.

\begin{figure}[!h]
\centering
\includegraphics[width=0.3\textwidth]{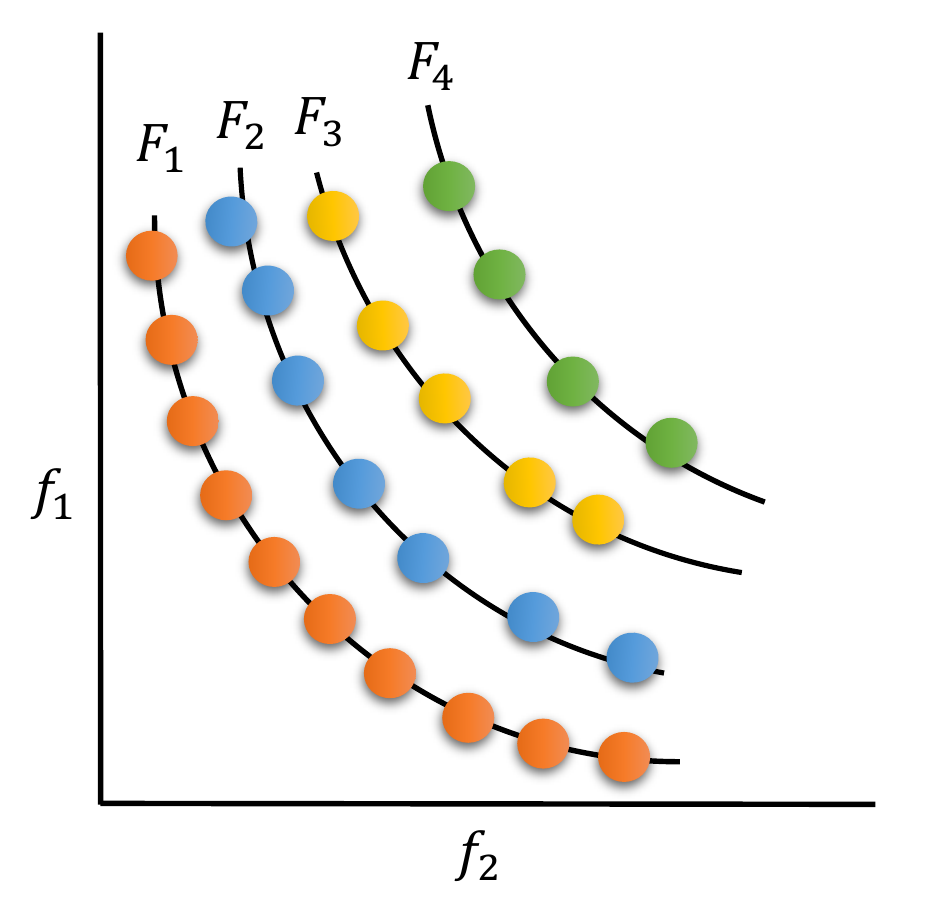} 
\caption{Example of a population $P$ partitioned into fronts $F_1,\ldots,F_n$ when minimizing objectives $f_1$ and $f_2$. $F_1$ is the Pareto front not dominated by any member of $P$.}
\label{pareto_front}
\end{figure}

In multi-objective optimization, optimal solutions are defined in terms of \emph{domination}.
A model $\mathbf{x}_1$ is said to dominate model $\mathbf{x}_2$,
if the following conditions are satisfied \cite{deb2001multi}:
 \begin{equation}
 \label{dominance_relation}
  \mathbf{x}_1\preceq \mathbf{x}_2 \ \mathbf{iff}
   \begin{cases}
   \forall i \in \{1,\dotsc ,M\} \; \quad f_i(\mathbf{x}_1)\leq f_i(\mathbf{x}_2)
    \\
    \exists j \in \{1,\dotsc ,M\} \quad f_j(\mathbf{x}_1)<f_j(\mathbf{x}_2)
   \end{cases}
   \end{equation}
The first condition states that the model $\mathbf{x}_1$ is no worse than $\mathbf{x}_2$ in all objectives $f_i$. The second condition states that the model $\mathbf{x}_1$ is strictly better than $\mathbf{x}_2$ in at least one objective. By using this concept, given the population of models $P$,
we can partition $P$ into $n$ sets called \emph{fronts} $F_1,\ldots , F_n$, such that $F_k$ dominates $F_l$ where $1 \leq k < l \leq n$ and the models within the same front do not dominate each other.
The so-called \emph{Pareto Front} or \emph{non-dominated set} $F_1$ includes models that are not dominated by any member of $P$. Essentially, using multi-objective optimization we efficiently find the best fitting models over a whole range of model complexities using a single coherent optimization approach. Figure \ref{pareto_front} provides a sketch.

\subsubsection{NSGA-{II}}
Non-dominated Sorting Genetic Algorithm {II} or NSGA-{II} \cite{deb2002fast} is a well-known multi-objective evolutionary algorithm (MOEA), still widely applied in various fields, such as image retrieval \cite{arevalillo2013hybrid}, reactive power planning \cite{hajabdollahi2012soft}, building design \cite{brownlee2015constrained}, and robot grippers \cite{saravanan2009evolutionary}. A characteristic feature is \emph{fast non-dominated sorting} which sorts models based on the concept of domination. With $M$ the number of objectives and $N$ the size of population, the time complexity has order $\mathcal{O}\left(MN^2\right)$, which is better than a na{\"\i}ve approach with
$\mathcal{O}\left( MN^3 \right)$.
 Another characteristic feature is \emph{crowding distance sorting} which is implemented to preserve the diversity among the solutions in the Pareto front. This feature sorts models based on the distance metric which explains the proximity of a model to other models.

\begin{figure}[!t]
\centering
\includegraphics[width=0.45\textwidth]{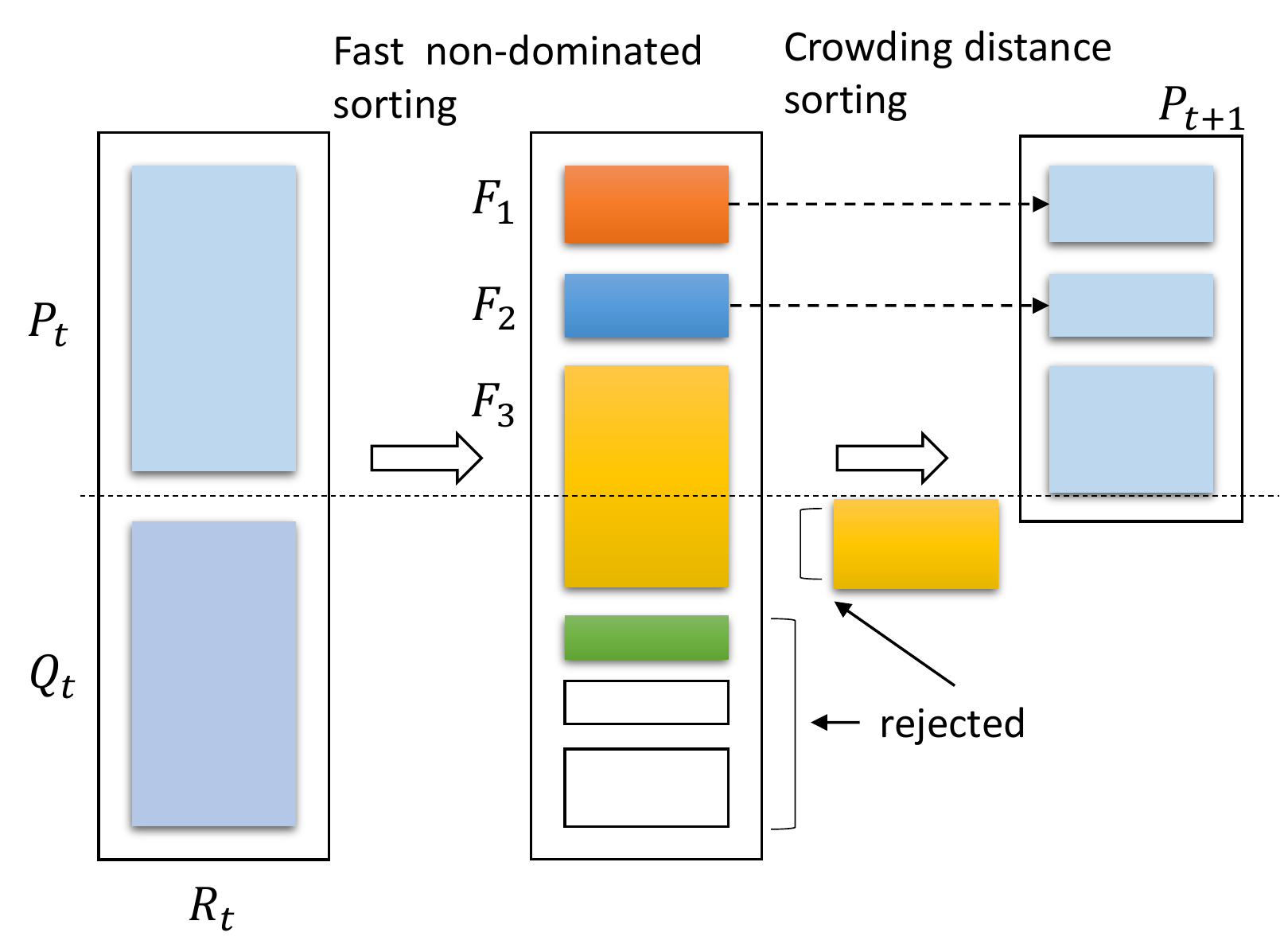}
\caption{Adopted from \cite{deb2002fast}. $P$ is the current population with size $N$ and is manipulated to make a new population $Q$. Both are combined, forming $R$, which will be sorted using fast non-dominated yielding a set of fronts $F$. Every member of front $F_n \in F$ will be assigned a so-called crowding distance in order to sort $F_k$. The first $N$ members of $F$ will be selected to be the next population $P$.}
\label{nsga}
\end{figure}

The iterative procedure of NSGA-{II} shown in Figure \ref{nsga} is a sequence of steps started by generating a population of solutions $P$ of size $N$. $P$ is then manipulated by genetic operators such as selection, crossover, and mutation, forming a new population $Q$ of size $N$. $P$ and $Q$ are then combined into population $R$ with size $2N$. After that $R$ is sorted using  fast non-dominated sorting, yielding a set of fronts $F$. In the next iteration each front in $F$ is sorted using the crowding distance sorting and the first $N$ members are used to generate a new population $P$. At $t=0$, $P$ is formed by creating $N$ random solutions sorted with fast non-dominated sorting.

\subsection{Stability Selection}
\label{subsec:stablePart}
Structure estimation is a notoriously difficult problem, both because of computational aspects (finding the optimal structure can be NP hard) and because of instability (small changes in the data can lead to completely different optimal structures). In this section we describe the method of \cite{meinshausen2010stability} for robust estimation of model structure based on subsampling in combination with selection algorithms. The method has been shown to yield finite sample family wise error control and improved structure estimates.

Let $\beta$ be a sparse $p$-dimensional vector which generally represents, for example, the coefficient vector in linear regression or the edges in a graph. In structure estimation the goal is to infer the set $S=\{k : \beta_k\not=0\}$ of non-zero components from noisy observations. Many methods tackle this problem by minimizing some loss function augmented with a regularization term to avoid overfitting. Usually the regularization term is parameterized by $\lambda\in\Lambda\subseteq\mathbb{R}^+$ and each $\lambda$ leads to an estimated structure $\hat{S}^\lambda\subseteq\{1,\ldots,p\}$. The objective is to determine $\lambda$ such that $\hat{S}^\lambda$ is identical to $S$ with high probability. To this end, \cite{meinshausen2010stability} introduces the concepts of \emph{selection probabilities} and \emph{stability paths}.

\begin{definition}[Selection probabilities]
Let $I$ be a subset of $\{1,\ldots,n\}$ of size $\lfloor n/2 \rfloor$ randomly drawn without replacement, $K\subseteq\{1,\ldots,p\}$, and
$\hat{S}^\lambda(I)$ be the selected set $\hat{S}^\lambda$ for subsample $I$. The probability of $K$ being in set
$\hat{S}^\lambda(I)$ is
\[
\hat{\Pi}^{\lambda}_K = P\left(K\subseteq\hat{S}^\lambda(I)\right)
\]
where the probability being is with respect to the random subsampling and possibly the construction of $\hat{S}^\lambda(I)$.
\end{definition}

\begin{definition}[Stability path]
For each variable $k=1,\ldots,p$ the \emph{stability path} is given by the selection probabilities $\{\hat{\Pi}_k^\lambda : \lambda\in\Lambda\}$.
\end{definition}

Furthermore, in stability selection we do not select a single element from the set of models $\{\hat{S}^\lambda : \lambda\in\Lambda\}$ as traditional methods do, but perturb the data many times and select structures that occur in a large fraction of selected sets. To this end, \cite{meinshausen2010stability} introduces the concept of \emph{stable variables}.

\begin{definition}[Stable variables] The set of stable variables is defined as
\[
\hat{S}^{stable} = \{k : \max\limits_{\lambda\in\Lambda} \hat{\Pi}^\lambda_k\geq\pi_{\mathrm{thr}} \}
\]
where $\pi_{\mathrm{thr}}$ is a cutoff with $0 < \pi_{\mathrm{thr}} < 1$.
\end{definition}

Variables with a high selection probability are kept whereas those with low selection probabilities are disregarded. The threshold $\pi_{\mathrm{thr}}$ is a tuning parameter but its influence is small and sensible values (e.g., $\pi_{\mathrm{thr}}\in (0.6,0.9)$) tend to give similar results.

\subsection{Model Equivalence}
There is one further subtlety that makes our approach for finding stable models (or sub-models) slightly more complicated than that in \cite{meinshausen2010stability}. If we find a particular model, we have to account for the fact that there may be different models that are observationally indistinguishable. Causal models represented by DAGs have their corresponding model equivalent classes, called \emph{Completed Partially Directed Acyclic Graph} (CPDAG). This means that every probability distribution derived from a model in a particular CPDAG, can also be derived by models belonging to the same CPDAG. In SEMs, these models are called covariance equivalent \cite{pearl2000causality}.

The characterization of equivalent structures is given by the following theorem \cite{verma1990equivalence}.
\begin{thm}
(Verma and Pearl, 1990) Two DAGs are equivalent if and only if they have the
same skeletons and the same v-structures.
\end{thm}
Furthermore, a directed edge $x\to y$ is compelled in $\mathcal{G}$ if for every DAG $\mathcal{G}'$ equivalent to $\mathcal{G}$, $x\to y$ exists in $\mathcal{G}$. For any edge $e$ in $G$, if $e$ is not compelled in $\mathcal{G}$, then $e$ is reversible. A CPDAG can be represented by a directed edge (arc) for every compelled edge and an undirected edge for every reversible edge \cite{chickering2002learning}.

Converting a model into a CPDAG allows one to observe the relations that hold among the variables. Arcs in a CPDAG indicate a cause-effect relation among variables since the same arc occurs in all members of the CPDAG. Undirected edges $A-B$ in a CPDAG indicate that some members of the CPDAG contain an arc $A\to B$ whereas other members contain an arc $B\to A$.

\section{Proposed method}
\label{sec:proposedMethod}
\subsection{The General Idea}
\label{subsec:generalIdea}
\begin{figure}[!b]
\centering
\includegraphics[width=0.5\textwidth]{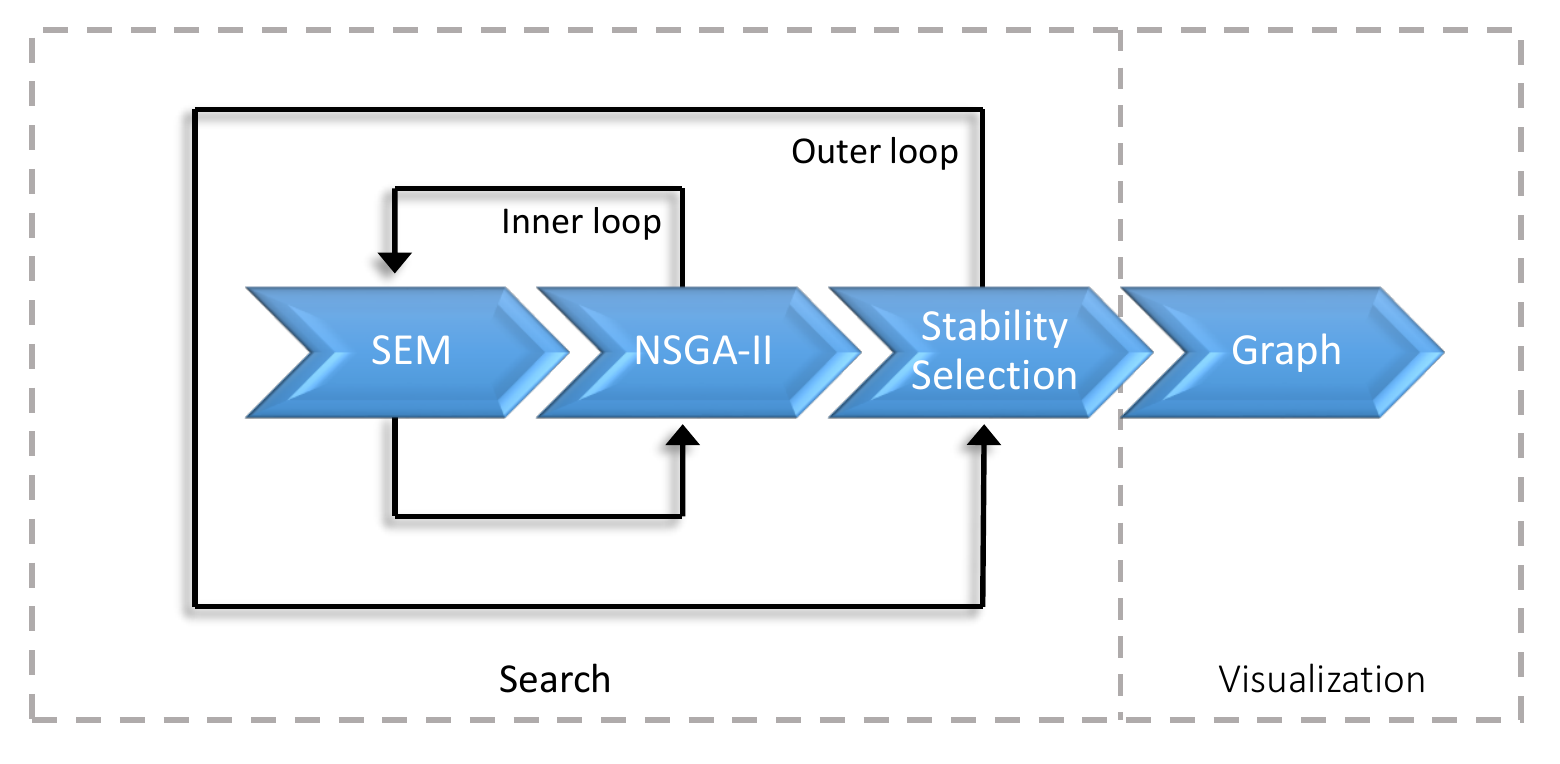}
\caption{The proposed method consists of two phases: \emph{search} and \emph{visualization}. The search phase is an iterative process using an \emph{outer loop} and \emph{inner loop} that combines SEM, \mbox{NSGA-II}, and stability selection, which
outputs all relevant edges and causal paths between two variables. The visualization phase displays the relevant relationships as a causal model.}
\label{fig_procedure}
\end{figure}

Our proposed method can be divided into two phases. The first phase is \emph{search} and the second phase is \emph{visualization}. In the search phase SEM and NSGA-II are synergically combined for exploratory search of the model space. As portrayed in Figure \ref{fig_procedure}, the \emph{inner loop} is an iterative process, searching over the model space and returns a Pareto front of models. The \emph{outer loop} is an iterative process that samples a different subset of the data in each iteration and at the end returns a number of Pareto fronts coming from those subsets.
Each model returned by the outer loop is transformed into a CPDAG which are then used to compute the \emph{edge stability graph} and the \emph{causal path stability graph}.

\begin{definition}
(Stability graphs) Let $A$ and $B$ be two variables and $G$ a multiset (or bag) of CPDAGS.  Let $G_c$ be the submultiset of $G$ containing all CPDAGS with complexity $c$. The edge stability for $A$ and $B$ at complexity $c$ is the number of models in $G_c$ for which there exists an edge between $A$ and $B$ (i.e., $A\rightarrow B$, $B\rightarrow A$, or $A-B$) divided by the total number of models in $G_c$. The causal path stability for $A$ to $B$ at complexity $c$ is the number of models in $G_c$ for which there is a directed path from $A$ to $B$ (of any length) divided by the total number of models in $G_c$. The terms edge stability graph and causal path stability graph are used to denote the corresponding measures for all variable pairs and all complexity levels.
\end{definition}

\begin{figure}[!t]
\centering
\mbox{
\subfloat[]
{
\includegraphics[width=0.45\textwidth]{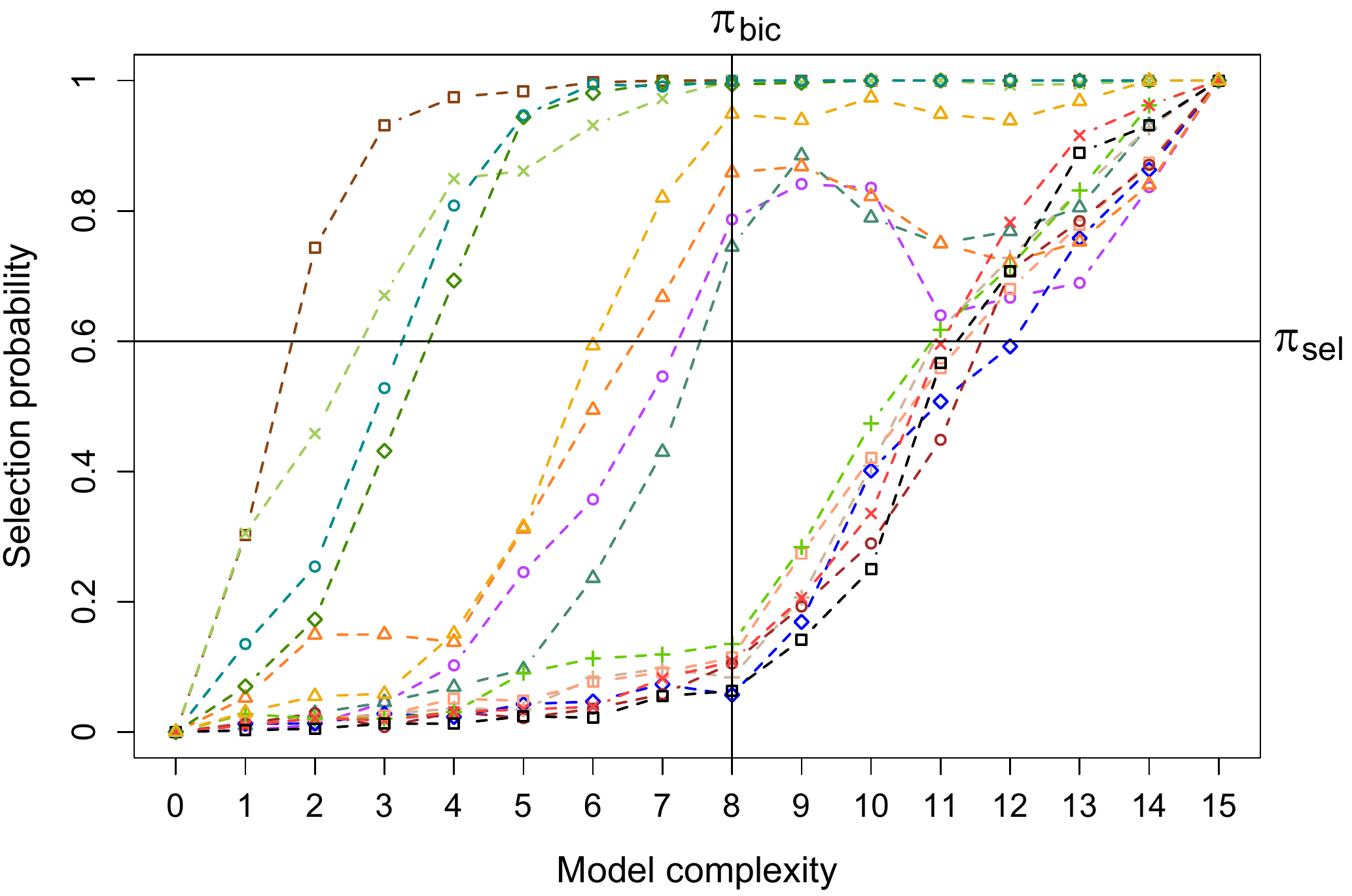}
\label{edgeStab}
}}
\mbox{
\subfloat[]
    {
    \includegraphics[width=0.45\textwidth]{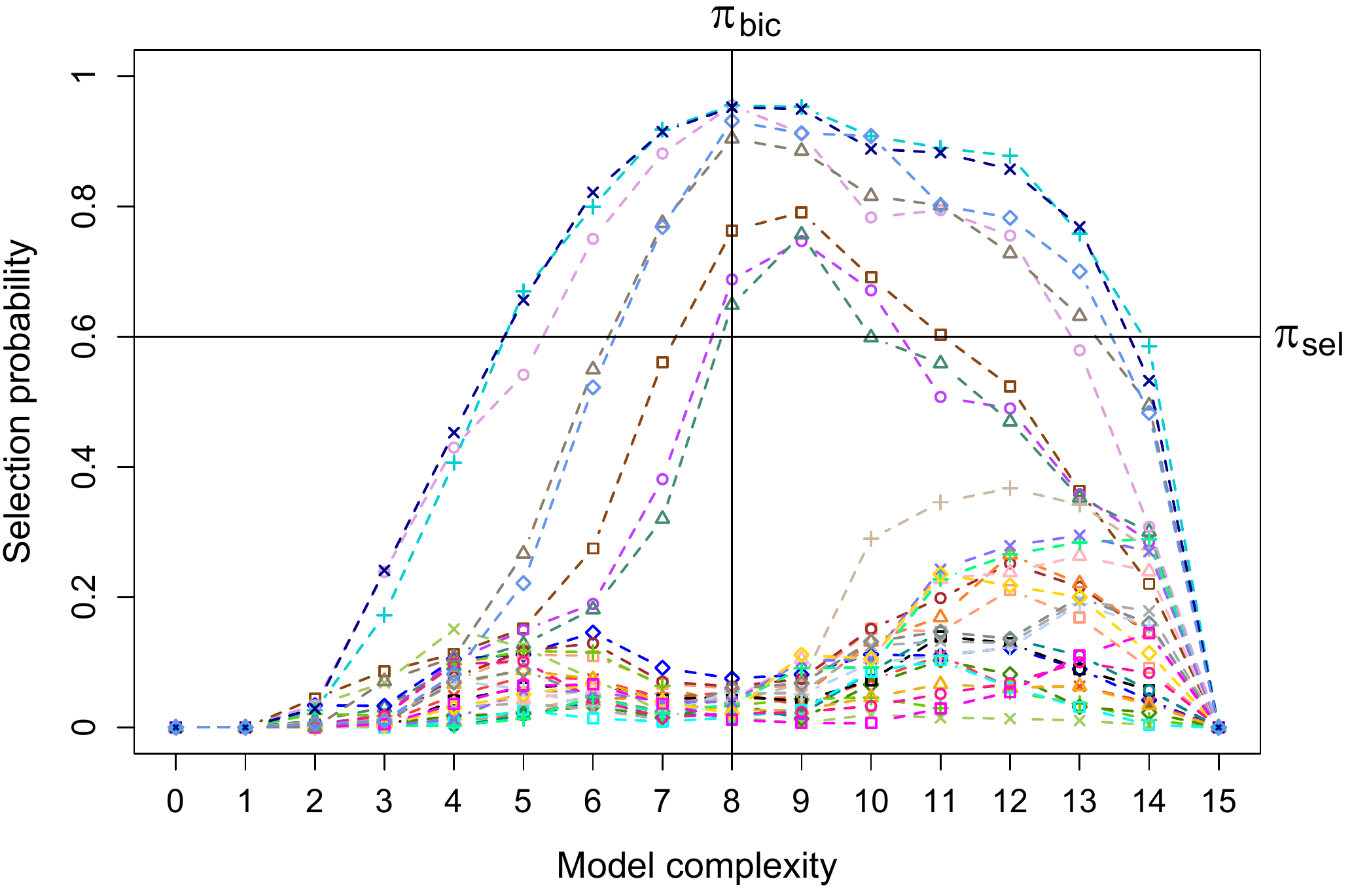}
        \label{causalStab}
    }
}
\caption{Example stability graphs from an artificial data set of $400$ instances with six continuous variables, without prior knowledge. (a) Edge stability graph. (b) Causal path stability graph. Each line in (a) represents an edge between a pair of variables and each line in (b) represents a causal path with any length from a variable to another variable. The threshold of selection probability, $\pi_{\mathrm{sel}}$, is set to $0.6$  and the threshold for model complexity, $\pi_{\mathrm{bic}}$, is chosen to minimize the average BIC score. See the main text for more details.}
\label{stabPlot}
\end{figure}

On top of the stability graphs we perform stability selection. In \cite{meinshausen2010stability}, stability selection is defined in terms of a regularization parameter $\lambda$. In our approach we do not have a regularization parameter and instead use model complexity (defined in Section \ref{sec:parameterSettings}) which is one of the objectives in our multi-objective optimization approach. We therefore define two thresholds. The first threshold is the boundary of selection probability $\pi_{\mathrm{sel}}$ and corresponds to $\pi_{\mathrm{thr}}$ in \cite{meinshausen2010stability}. For example, setting \mbox{$\pi_{\mathrm{sel}}=0.6$} means that all causal relationships with edge stability or causal path stability (Figure~\ref{stabPlot}) above this threshold are considered \emph{stable}. The second threshold is the boundary of complexity $\pi_{\mathrm{bic}}$, which is used to control overfitting and corresponds to minimal $\lambda$ in \cite{meinshausen2010stability}. We set $\pi_{\mathrm{bic}}$ to the level of model complexity at which the minimum average \emph{Bayesian Information Criterion} (BIC) score is found. For example, $\pi_{\mathrm{bic}}=7$ means that all causal relationships with an edge stability or a causal path stability lower than this threshold (Figure~\ref{stabPlot}) are considered \emph{parsimonious}.
Causal relationships that intersect with the top-left region are considered both stable and parsimonious and called \emph{relevant}.

In the visualization phase we combine the stability graphs into a graph with nodes and edges. This is done by adding the relevant edges and orienting them using background knowledge (if any, see Section \ref{subsec:dagToCpdag}) and the relevant causal paths. In addition we annotate each edge with the highest selection probability it has across different model complexities in the top-left region of the edge stability. This visualization eases interpretation but the stability graphs are considered to be the main outcome of our approach.

\subsection{Constrained SEM}
\label{subsec:dagToCpdag}
In practise, one often has prior knowledge about the domain, for example, that $A$ does not cause $B$ directly, denoted by $A \not\to B$. The method proposed here can include such prior knowledge, extending previous work \cite{rahmadi2014}, since this translates to a DAG with no directed edge from $A$ to $B$.\footnote{This still allows for directed edges from $B$ to $A$ or indirect relations from $A$ to $B$.}

Model specifications should comply with any prior knowledge when performing specification search and when measuring the edge and causal path stability. When DAGs are converted into CPDAGs in the outer loop, a constraint  $A \not\to B$ may be violated since arcs $B \to A$ in the DAG may be converted into undirected (reversible) edges $A-B$ in the CPDAG. In order to preserve constraints we therefore extended the efficient DAG-TO-CPDAG algorithm of \cite{chickering2002learning} which runs in time $\mathcal{O}\left(|E|\right)$ given a DAG $G=(V,E)$.
\begin{figure*}
\noindent\fbox{%
	 \begin{minipage}{\dimexpr\linewidth-2\fboxsep-2\fboxrule\relax}
  \begin{algorithmic}[1]
    \Function{$\mathrm{consDag2Cpdag}$}{DAG $G,$ constraint $\mathcal{C}$}
      \State $E' \gets \mathrm{orderEdges}(G)$
      \For{every constraint $c \in \mathcal{C}$}
        \State get $e \in E'$ that matches c
        \State label $e$ with ``compelled" in the direction consistent with $c$
      \EndFor
      \State return $G' \leftarrow \mathrm{labelEdges} (G,E')$ \Comment{label remaining edges using  \cite{chickering2002learning}}
    \EndFunction
  \end{algorithmic}
  \end{minipage}%
	}
  \caption{The constrained DAG-TO-CPDAG algorithm returns a CPDAG which is consistent with the added prior knowledge and extends \cite{chickering2002learning}. The algorithm first labels the edges that match the constraints with "compelled" and then labels the remaining edges with "reversible" or "compelled" using  \cite{chickering2002learning}.}
  \label{mod_chickering2}
\end{figure*}

Figure~\ref{mod_chickering2} provides pseudocode for the constrained DAG to CPDAG algorithm.
Line 2 produces a total ordering $E'$ over the edges in DAG $G$. Lines 3-6 impose an arc upon the edges that match the constraints. Finally, Line 7 uses \cite{chickering2002learning} to label the remaining edges $E \setminus{E'}$ in $G$ with ``compelled'' or ``reversible'' and returns the constrained CPDAG $G'$.

A DAG without edges will always be transformed into a CPDAG without edges. A fully connected DAG without constraints will be transformed into a CPDAG with only undirected edges. However, if background knowledge is added, a fully connected DAG will be transformed into a CPDAG in which the edges corresponding to the background knowledge are directed. From these observations it follows that in the edge stability graph all paths start with a selection probability of 0 and end up in a selection probability of 1. In the causal path stability graph when no prior knowledge has been added all paths start with a selection probability of 0 and end up in a selection probability of 0. However, when prior knowledge is added some of the paths may end up in a selection probability of 1 because of the added constraints.

\subsection{Stable Specification Search Algorithm}
Figure~\ref{proposedProc} provides \mbox{pseudocode} for our approach (cf. Figure~\ref{fig_procedure}).
Lines~\mbox{3-18} represent the outer loop, Lines 6-16 represent the inner loops, Lines~19-21 compute stability graphs.
\begin{figure*}[!t]
  \label{proposed_method_alg}
  \noindent\fbox{%
	 \begin{minipage}{\dimexpr\linewidth-2\fboxsep-2\fboxrule\relax}
  	\begin{algorithmic}[1]
    \Procedure{$\mathrm{stableSpecificationSearch}$}{data set $D,$ constraint $\mathcal{C}$}
      \State $H\gets ()$ \Comment{initialize}
      \For{$j\gets 0,\dotsc ,J-1$}\Comment{$J$ is number of outer loop iterations}
        \State $T\gets$ subset of $D$ with size $\lfloor |D|/2\rfloor$ without replacement
        \State $F_1 \gets ()$ \Comment{initialize Pareto fronts to empty list}
        \For{$i\gets 0,\dotsc ,I-1$}\Comment{$I$ is number of inner loop iterations}
        	\If{$i=0$}
        		\State $P \gets N$ random DAGs consistent with $\mathcal{C}$
        		\State $P \gets$ $\mathrm{fastNonDominatedSort(P)}$
        		\Else
        			\State $P \gets \mathrm{crowdingDistanceSort(F)}$ \Comment{draw the first $N$ models}
        	\EndIf
        	\State $Q \gets$ make population from $P$
        	\State $F\gets \mathrm{fastNonDominatedSort(P^\frown Q )}$
        	\State $F_1\gets$ pareto front of $F$ and $F_1$
        \EndFor
        \State $H \gets H^\frown F_1$ \Comment{concatenation}
      \EndFor
      \State $G \gets \mathrm{consDag2Cpdag}(H, \mathcal{C})$
      \State edges $\gets$ edge stability of $G$
  	  \State paths $\gets$ path stability of $G$
    \EndProcedure
  \end{algorithmic}
  \end{minipage}%
	}
    \caption{Stable specification search consists of an outer and an inner loop. The outer loop samples a subset of the data, and for every subset, the inner loop searches for the Pareto front by applying NSGA-II. The Pareto fronts are converted into constrained CPDAGs which are then used to compute the edge and causal path stability graph.}
    \label{proposedProc}
\end{figure*}

An inner loop \mbox{(Lines 6-16)} starts by forming a population $P$ of size $N$, initially at random, or else from a previous population using crowding distance sorting \mbox{(Lines 7-12)}. Models are represented with a binary vector $\bf{y}$ with $y_i\in\{0,1\}$ denoting the existence of some arc $A\rightarrow B$.
\mbox{Line 13} forms a new population $Q$ by manipulating $P$ using binary tournament selection, one-point crossover, and one-bit flip mutation, which are compatible with a binary representation. The selection scheme selects $N$ times two models from $P$ and places the best model (i.e., lowest front or else smallest crowding distance) in a mating pool $M_{pool}$. One-point crossover takes two models from $M_{pool}$ and swaps the data after the crossover point (the middle). One-bit flip mutation flips each bit according to a predetermined rate.
\mbox{Line 14} combines $P$ and $Q$ and sorts them using fast non-dominated sorting. Line 15 updates the Pareto front in $F_1$.

An outer loop \mbox{(Lines 3-18)} randomly samples a subset $T$ from $D$ with size $\lfloor |D|/2 \rfloor$ \mbox{(Line 4)}, runs the inner loop $I$ times to obtain a Pareto front \mbox{(Lines 6-16)}, and stores it in $H$ \mbox{(Line 17)}. After $J$ iterations, $H$ contains $J$ Pareto fronts.

\mbox{Lines 19-21} convert the $J$ Pareto fronts in $H$ from DAGs into CPDAGs using the algorithm in Figure~\ref{proposedProc} and then computes the edge and causal path stability graphs. The stability graphs are considered to be the main outcome of our approach, but can also be visualized as a graph with nodes and edges.

\section{Experimental Study}
\label{sec:experimental}
We implemented the stable specification search as an R package named \texttt{stablespec}.
The package is publicly available at the Comprehensive \textsf{R} Archive Network (CRAN)\footnote{\url{https://cran.r-project.org/web/packages/stablespec/index.html}}, so it can be installed directly, e.g., from \textsf{R} console by typing \texttt{install.package("stablespec")} or from RStudio by using feature to install package. We also included a package documentation as a brief tutorial of using the functions. All experiments were run on an Intel Xeon E7-4870 v2 Processor 2.3 GHz, 15 Core, 96 of 32GB LRDIMM.

\subsection{Parameter Settings}
\label{sec:parameterSettings}
For all experiments, we employed the same set of NSGA-{II} parameters and stability \mbox{thresholds}. We had $100$ iterations in the outer loop, and in each iteration we drew a subsample with size $\lfloor |D|/2 \rfloor$.
We did not do a comprehensive parameter tuning for NSGA-{II}, instead, we followed guidelines provided in \cite{grefenstette1986optimization}.
The parameters were set as follows: the number of generations (inner loop) was $20$, the size of the population $P$ was $100$, the crossover rate was $0.85$, the mutation rate was $0.075$ and with binary tournament selection.

We score models using the \emph{chi-square} $\chi^2$ and the \emph{model complexity}. The $\chi^2$ is considered the original fit index in SEM and measures whether the model-implied covariance matrix is close enough to the sample covariance matrix \cite{kline2011principles}.

The model complexity represents how many predicted parameters the model contains. Assuming that variances of parameters are always predicted, the maximum model complexity with $n$ variables is given by $n(n-1)/2$.

When using multi-objective optimization we minimize both the $\chi^2$ and model complexity objectives.
These two objectives are, however, conflicting with each other. For example, minimizing the model complexity typically means compromising the data fit.

\subsection{Application to Simulated Data}
\subsubsection{Data Generation}
\begin{figure}[!t]
\centering
\includegraphics[width=0.35\textwidth]{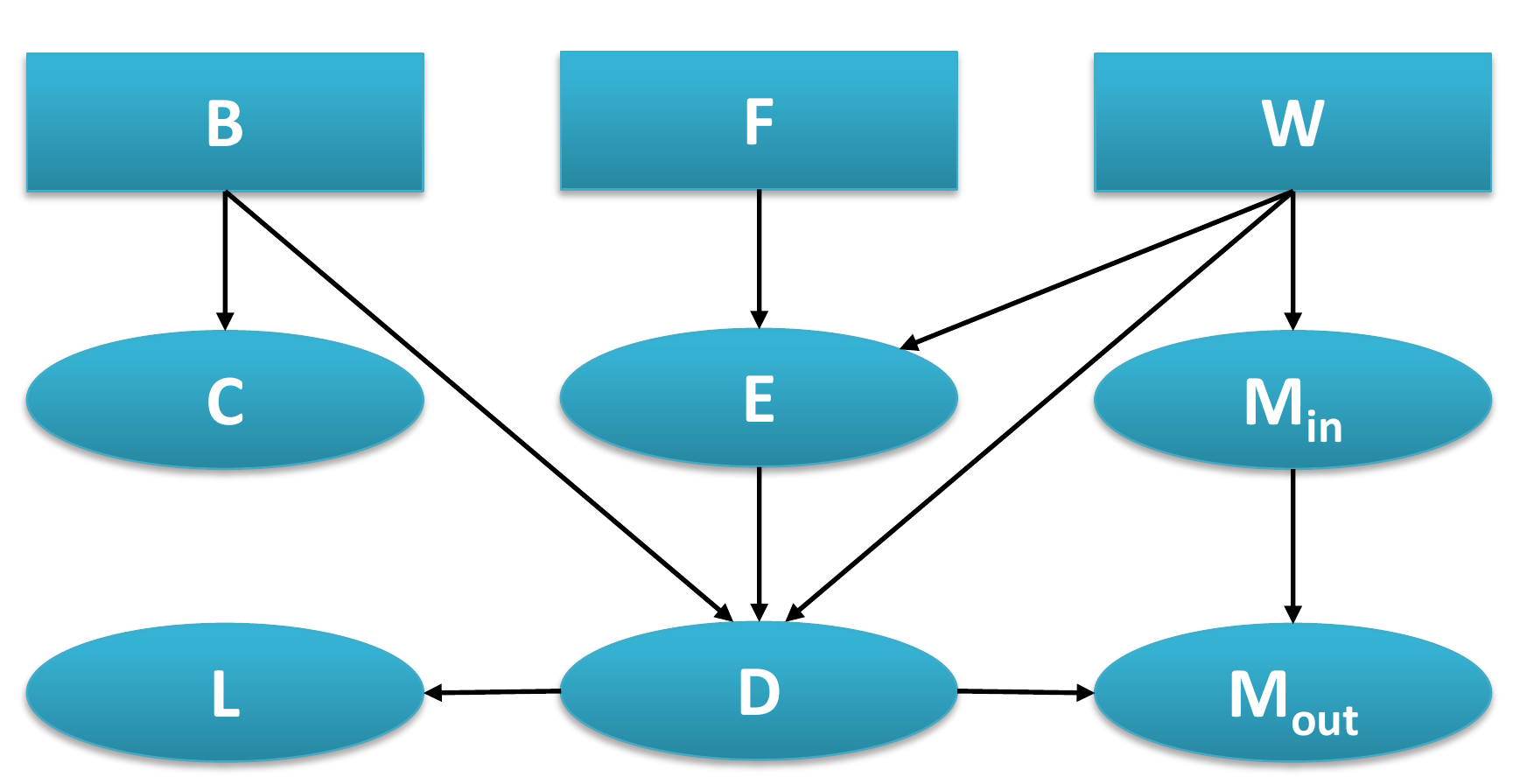}
\caption{The Waste Incinerator network. Rectangular nodes represent discrete variables, oval nodes represent continuous variables, and arcs represent direct causal relations.}
\label{waste_model}
\end{figure}

In this experiment we generated data using the Waste Incinerator network in Figure~\ref{waste_model}, which is a model of waste emissions from an incinerator plant \cite{lauritzen1992propagation}. This model contains both discrete and continuous random variables, with $\mathrm{B}$ the waste burning regimen, $\mathrm{W}$ the compositional differences in incoming waste, $\mathrm{C}$ the concentration of CO2, $\mathrm{F}$ the filter state, $\mathrm{E}$ the filter efficiency, $\mathrm{L}$ the light penetrability, $\mathrm{D}$ the emission of dust, $\mathrm{M_{in}}$ the metals in waste, and $\mathrm{M_{out}}$ the metals emission. Following~\cite{rhemtulla2012can}, we treat all discrete variables as continuous. We added prior knowledge that none of the variables directly cause the filter state.
We generated $10$ data sets containing $400$ samples from this network using the BNT toolbox with the default parameter setting as described in \cite{murphy2001bayes}.

\subsubsection{Performance Measure}
We compared the stable specification search with GES (score-based method), PC-stable, and CPC (both constrained-based methods). Our method intrinsically subdivides the data
in a number of subsets, here $50$ of size $200$ samples, and then runs the multi-objective optimization to obtain $50$ Pareto fronts (see Section \ref{sec:multi-objective}).
For a fair comparison, for each algorithm we consider subsampling (e.g., \cite{ramsey2010bootstrapping}), giving each method $50$ subsets. For every subset, each algorithm returns a CPDAG from which we can derive the edges and causal paths.

Since the true model of the Waste Incinerator data is known, we can measure the performance of both methods by means of the \emph{Receiver Operating Characteristic} (ROC) curve \cite{fawcett2004roc}.
The \emph{True Positive Rate} (TPR) and the \emph{False Positive Rate} (FPR) are computed with respect to the CPDAG of the true model. For example, in the case of causal path stability, a true positive means that a causal path with any length obtained through our approach or the PC algorithm is actually present in the CPDAG of the true model. By increasing the threshold $\pi_{\mathrm{sel}}$, we increase the TPR at the expense of the FPR. In addition, we conducted three significance tests to compare the ROC curves. The first test \cite{delong1988comparing} compares the \emph{Area Under the Curve} (AUC) of the ROC curves based on the theory of U-statistics. The second test \cite{robin2011proc}, a modification of \cite{hanley1983method}, compares the AUC of ROC curves that are generated from bootstrap replicates. The third test \cite{venkatraman1996distribution} compares the actual ROC curves by evaluating the absolute difference. The null hypothesis is that the AUC of the ROC curves of our method and the PC algorithm are equivalent.

We repeated the above procedure $10$ times on different Waste Incinerator data sets and computed the ROC curves using two different schemes: averaging and individual. In the averaging scheme, the ROC curves are computed based on the average edge and causal path stability from different data sets. We conducted statistical significance tests on these average ROC curves.  Conversely, in the individual scheme the ROC curves are computed directly from the edge and causal path stability on each data set. We conducted individual statistical significance tests on the ROC curves for each data set and then used Fisher's method, as described in \cite{fisher1925statistical, 10.2307/2681650}, to  combine these tests into a single test statistic. Both schemes are intended to show empirically and comprehensively how robust the results of each algorithm are across changes in the data.

\subsubsection{Discussion of Waste Incinerator Result}
\label{subsection:discussion_waste}
\begin{figure}[!t]
\centering
\mbox{
    \subfloat[]{
        \includegraphics[width=0.45\textwidth]{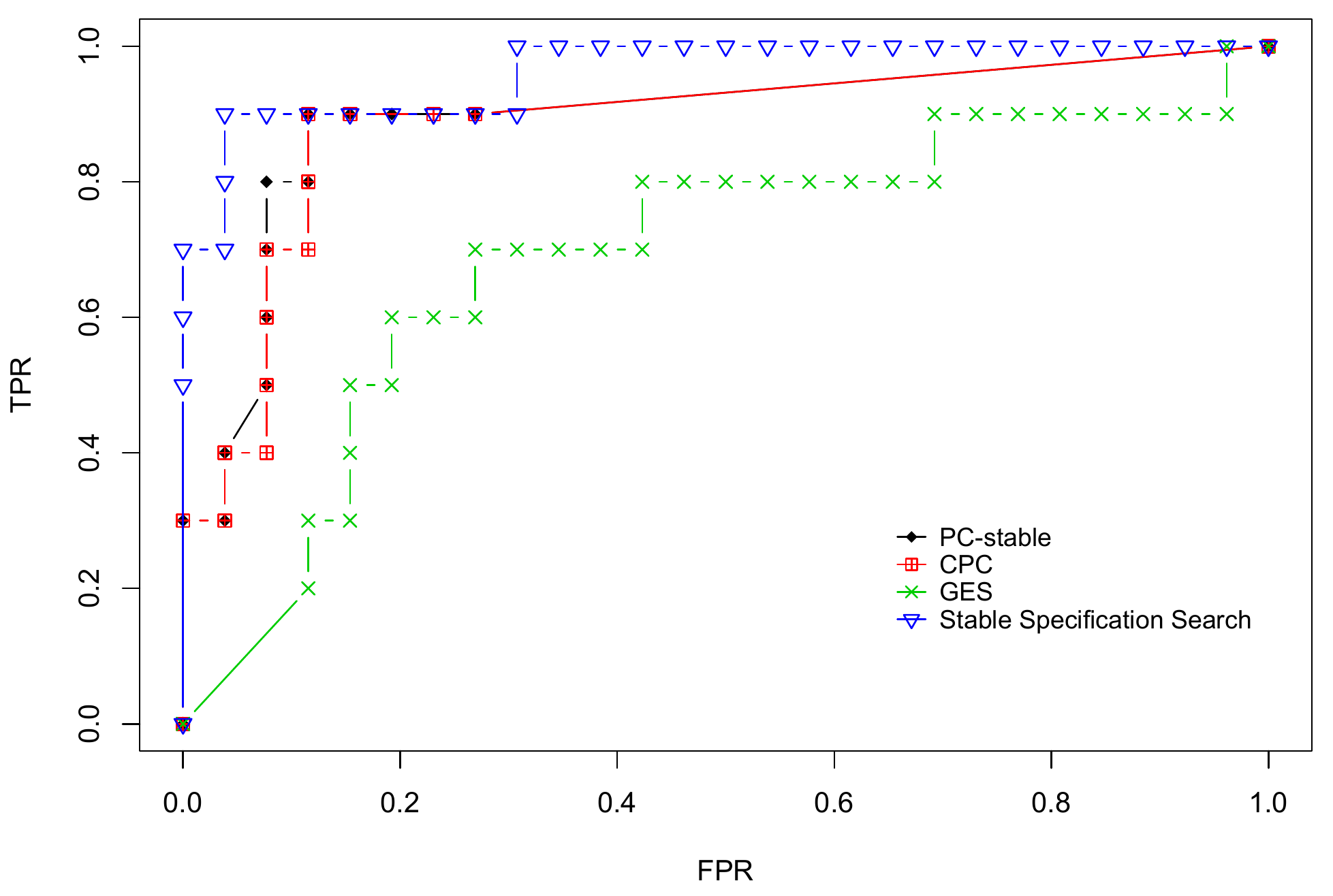}
        \label{edgeROC}
        }}
    \mbox{
    \subfloat[]{
        \includegraphics[width=0.45\textwidth]{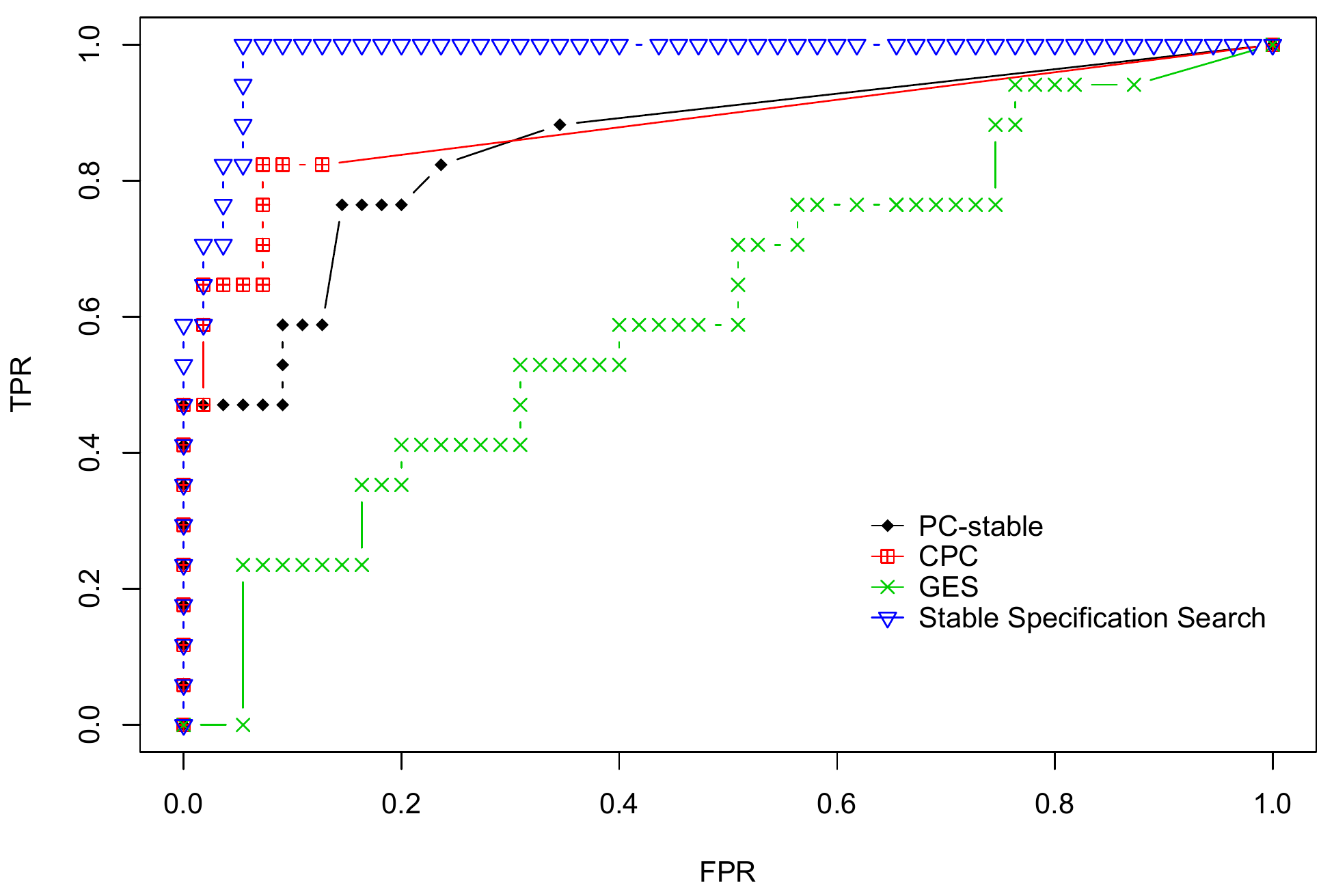}
        \label{causalROC}
        }}
\caption{ROC curves for (a) the edge stability and (b) the causal path stability, for different values of $\pi_{\mathrm{sel}}$ in the range of $[0,1]$. In (a), the AUCs are 0.96 (stable specification search), 0.89 (PC-stable), 0.88 (CPC), and 0.69 (GES). In (b), The AUCs are 0.98 (stable specification search), 0.85 (PC-stable), 0.88 (CPC), and 0.61 (GES).}
\label{rocWaste}
\end{figure}

Figure~\ref{rocWaste} shows the ROC curves for (a) the edge stability and (b) the causal path stability from the averaging scheme.
The corresponding AUCs for edge stability are $0.96$ (stable specification search), $0.89$ (PC-stable), $0.88$ (CPC), and $0.69$ (GES). The AUCs for causal path stability are $0.98$ (stable specification search), $0.85$ (PC-stable), $0.88$ (CPC), and $0.61$ (GES).

Table~\ref{tableROC_comp} lists the results of the significance tests for both the averaging and individual schemes. The ROC and AUC for the edge stability are comparable with PC-stable and CPC (\emph{p}-value $> 0.1$), but always significant (\emph{p}-value $<0.01$) compared with GES. The ROC and AUC for the causal path stability compared with CPC are marginally significant (\emph{p}-value $<0.1$) using the averaging scheme, but significant using the individual scheme (\emph{p}-value $<0.01$); compared with PC-stable significant (\emph{p}-value $<0.05$) using the averaging scheme, but highly significant using the individual scheme (\emph{p}-value $<10^{-5}$); compared with GES highly significant using both schemes (\emph{p}-value $<10^{-5}$).
To conclude, we show that the stable specification search obtains at least comparable performance as, but often significant improvement over alternative approaches, especially in obtaining the causal relations.

\label{subsection:discussion_waste}

\begin{table*}[!t]
\centering
\normalsize
\caption{Table of \emph{p}-values from comparisons between stable specification search and alternative approaches. For each significance test, we compared the ROC of the edge (Edge) and causal path (Causal) stability on both averaging (Ave.) and individual (Ind.) schemes.}
\begin{tabular}{|c|c|c|c|c|c|c|c|}
\cline{3-8}
\multicolumn{2}{l}{} &  \multicolumn{2}{|c|}{GES} & \multicolumn{2}{c|}{PC-stable} & \multicolumn{2}{c|}{CPC}\\
\hline
\multicolumn{2}{|c|}{Significance test} & Ave. & Ind. & Ave. & Ind. & Ave. & Ind. \\
\hline
{DeLong \cite{delong1988comparing}} & {Edge} & $0.003$ & $<10^{-5}$ & $0.317$ & $0.175$ & $0.284$ & $0.131$\\ {} & {Causal} & $<10^{-5}$ & $<10^{-5}$ & $0.027$ & $<10^{-5}$ & $0.073$ & $<10^{-5}$ \\ \hline
{Bootstrap \cite{robin2011proc}} & {Edge} & $0.003$ & $<10^{-5}$ & $0.296$ & $0.135$ & $0.261$ & $0.098$ \\
{} & {Causal} & $<10^{-5}$ & $<10^{-5}$ & $0.022$ & $<10^{-5}$ & $0.064$ & $<10^{-5}$ \\ \hline
{Venkatraman \cite{venkatraman1996distribution}} & {Edge} & $0.004$ & $<10^{-5}$ & $0.591$ & $0.684$ & $0.539$ & $0.592$ \\
{} & {Causal} & $<10^{-5}$ & $<10^{-5}$ & $0.023$ & $<10^{-5}$ & $0.096$ & $0.005$ \\ \hline
\end{tabular}
\label{tableROC_comp}
\end{table*}

\subsection{Application to Real-world Data}
This section describes the results of applying our proposed method on two real-world data sets. Both of them are about particular diseases, for which the underlying causal relationships are often not clear. Revealing such causal relationships can lead to the development of (new) dedicated treatments and medications. Here, we consider data on \emph{Attention Deficit Hyperactivity Disorder} (ADHD) and \emph{Chronic Fatigue Syndrome} (CFS).

\subsubsection{Performance Measure}
Since the true model is unknown we measure the performance of our method using the edge stability and causal path stability graphs. We set the thresholds to $\pi_{\mathrm{sel}} = 0.6$ and $\pi_{\mathrm{bic}}$ to the minimum average of BIC scores.
The relevant causal relations are those which occur in the top-left region (see Figure~\ref{stabPlot} as example).
We compare the stability graphs to studies reported in the literature.

\subsubsection{Application to CFS}
In this experiment we consider a data set about \emph{Chronic Fatigue Syndrome} (CFS) of $183$ subjects \cite{heins2013process}. Originally the data comes from a longitudinal study with five time slices, but in this paper, we focus only on one time slice representing the subjects after the first treatment.

The data set contains six discrete variables; $\mathrm{fatigue}$ severity assessed with the subscale fatigue severity of the Checklist Individual Strength (CIS), the sense of $\mathrm{control}$ over fatigue assessed with the \emph{self-efficacy scale} (SES), $\mathrm{focusing}$ on symptoms measured with the \emph{Illness Management Questionnaire}, the objective activity of the patient measured using an \emph{actometer} ($\mathrm{oActivity}$), the subject's perceived activity measured with the subscale activity of the CIS ($\mathrm{pActivity}$), and physical $\mathrm{functioning}$ measured with subscale physical functioning of the \emph{medical outcomes survey} (SF36). We refer to the original paper \cite{heins2013process}, for a detailed description of the questionnaires used and the actometer. Missing values were imputed using an imputation method \emph{Expectation Maximization} implemented in SPSS \cite{spss}. As all of the variables have large scales, e.g., in the range between $0$ to $155$, we treat them as continuous variables. We added prior knowledge that the variable $\mathrm{fatigue}$ does not cause any of the other variables directly.

\begin{figure}[!t]
\centering
\mbox{
    \subfloat[]{
        \includegraphics[width=0.45\textwidth]{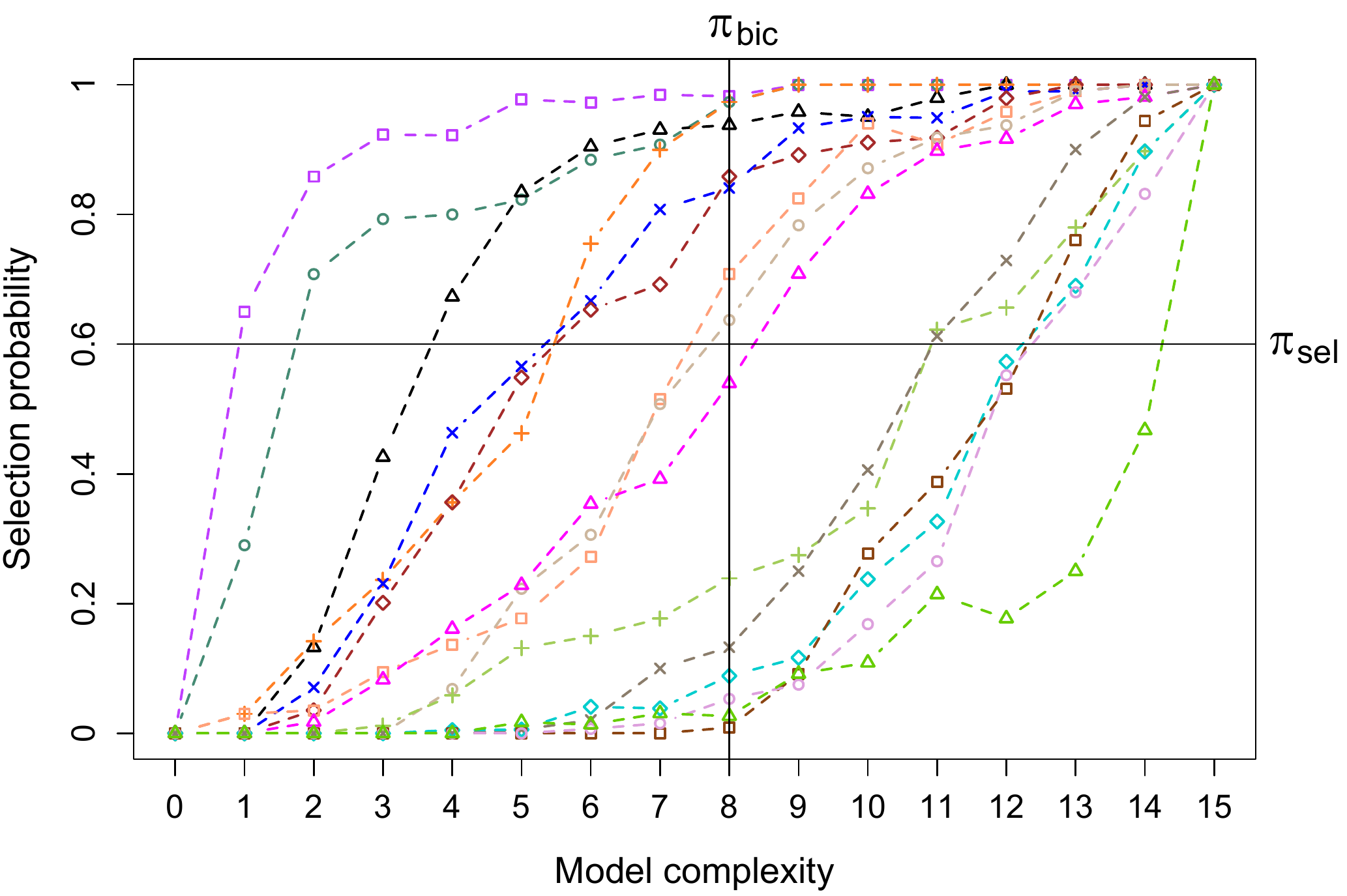}
        \label{edgeStabCFS}
        }
    }
\mbox{
    \subfloat[]{
        \includegraphics[width=0.45\textwidth]{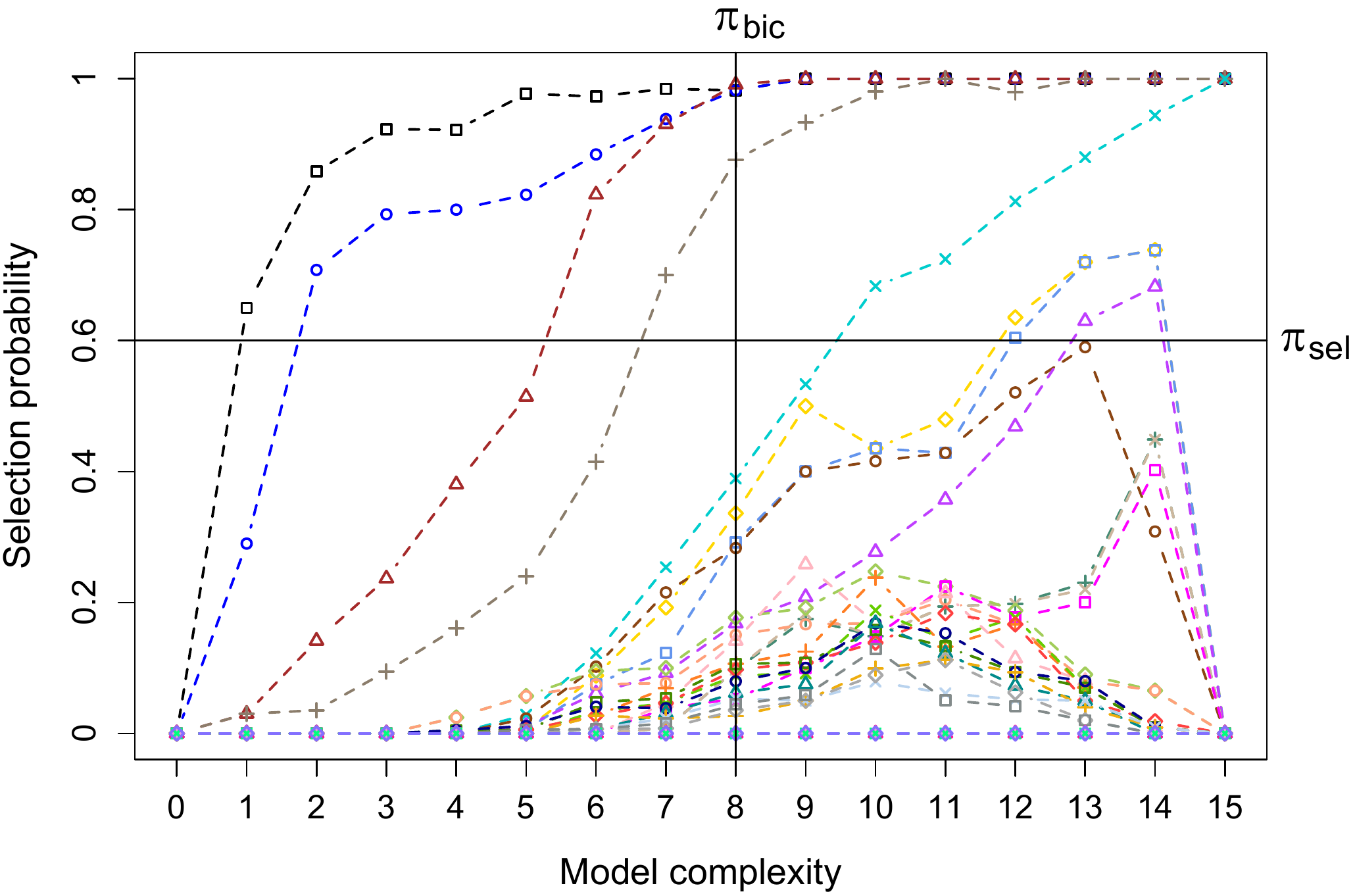}
        \label{causalStabCFS}
        }
    }
\caption{The stability graphs for CFS together with $\pi_{\mathrm{sel}}$ and $\pi_{\mathrm{bic}}$, yielding four regions. The top-left region is the area containing the relevant causal relations. (a) The edge stability graph showing eight relevant edges. (b) The causal path stability graph showing four relevant causal paths. See Tables~\ref{tableRelevantEdgeCFS} and~\ref{tableRelevantCausalCFS} in Appendix \ref{sec:appendixA} for more detail.}
\label{stabCFS}
\end{figure}

The total computation time for one subset was around $5.5$ minutes. Figure~\ref{stabCFS} shows that eight relevant edges were found. These edges are between $\mathrm{pActivity}$ and $\mathrm{fatigue}$, $\mathrm{focusing}$ and $\mathrm{fatigue}$, $\mathrm{functioning}$ and $\mathrm{fatigue}$, $\mathrm{control}$ and $\mathrm{fatigue}$, $\mathrm{pActivity}$ and $\mathrm{focusing}$, $\mathrm{pActivity}$ and $\mathrm{oActivity}$, $\mathrm{focusing}$  and $\mathrm{control}$, and $\mathrm{functioning}$ and $\mathrm{control}$.

Figure \ref{stabCFS} shows that four relevant causal paths were found. These causal paths are: $\mathrm{pActivity}$ to $\mathrm{fatigue}$, $\mathrm{control}$ to $\mathrm{fatigue}$, $\mathrm{functioning}$ to $\mathrm{fatigue}$, and $\mathrm{focusing}$ to $\mathrm{fatigue}$.

\begin{figure}[!t]
\centering
\includegraphics[width=0.45\textwidth]{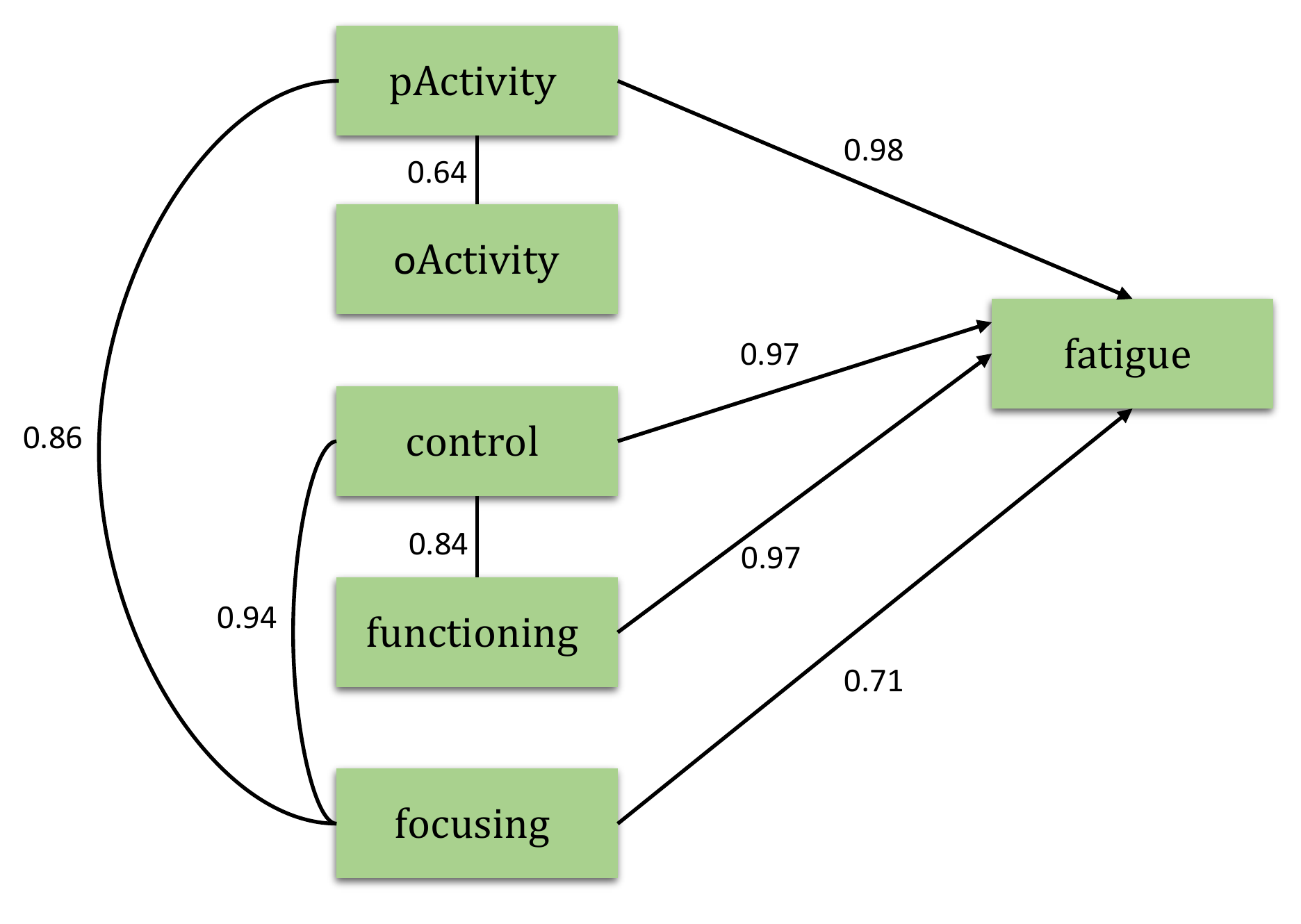} 
\caption{The inferred model of CFS by combining the edge stability and causal path stability graphs. Each edge has a reliability score which is the highest selection probability in the top-left region of the edge stability graph.}
\label{CFS_model}
\end{figure}

The stability graphs can be combined into a model as follows. First, the nodes are connected according to the eight relevant edges obtained. Second, the edges are oriented according to the background knowledge added. The fact that the variable $\mathrm{fatigue}$ does not directly cause any other variable results in four directed edges, which, in this case, correspond exactly to the relevant causal paths obtained. The inferred model is shown in Figure~\ref{CFS_model}.

A (direct) causal path $X\rightarrow Y$ in Figure~\ref{CFS_model} indicates that a change in variable $X$ causes a change in variable $Y$. All variables except for objective activity were found to be direct causes for fatigue severity, which are corroborated by literature studies. In \cite{vercoulen1998persistence}, changes in physical activity, sense of control, and focus on symptoms measured, were shown to result in changes in fatigue. In \cite{wiborg2012towards}, changes in perceived activity, sense of control, and physical functioning were shown to result in changes in fatigue. In \cite{heins2013process}, an increase in sense of control, perceived activity, and self-reported physical functioning, as well as a decrease in focusing on symptoms resulted in a decrease of fatigue, whereas changes in objective activity did not result in any change in fatigue.

\subsubsection{Application to ADHD}
\begin{figure}[!t]
\centering
\mbox{
\subfloat[]{
    \includegraphics[width=0.45\textwidth]{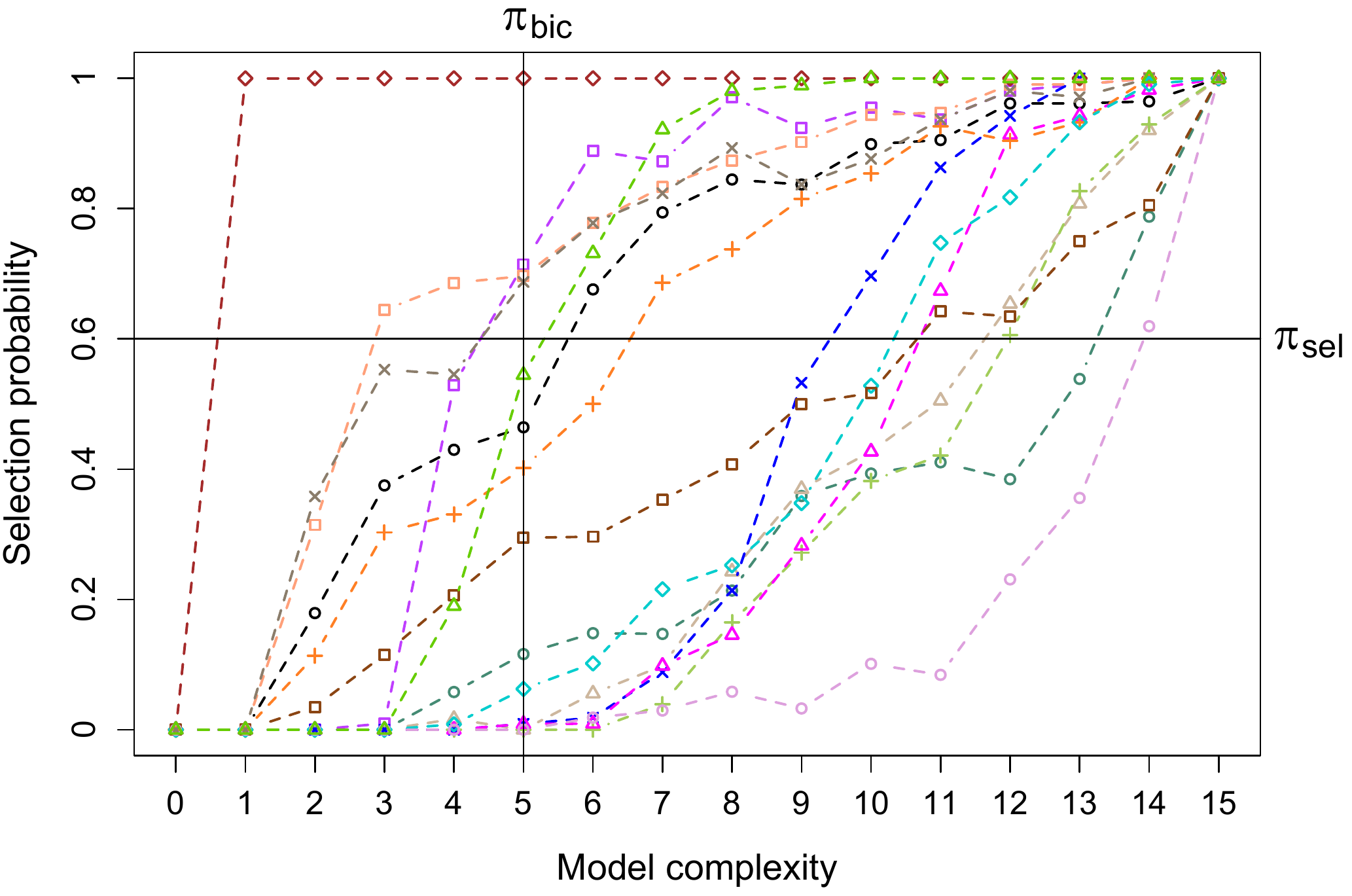}
    \label{edgeStabADHD}
}}
\mbox{
\subfloat[]{
    \includegraphics[width=0.45\textwidth]{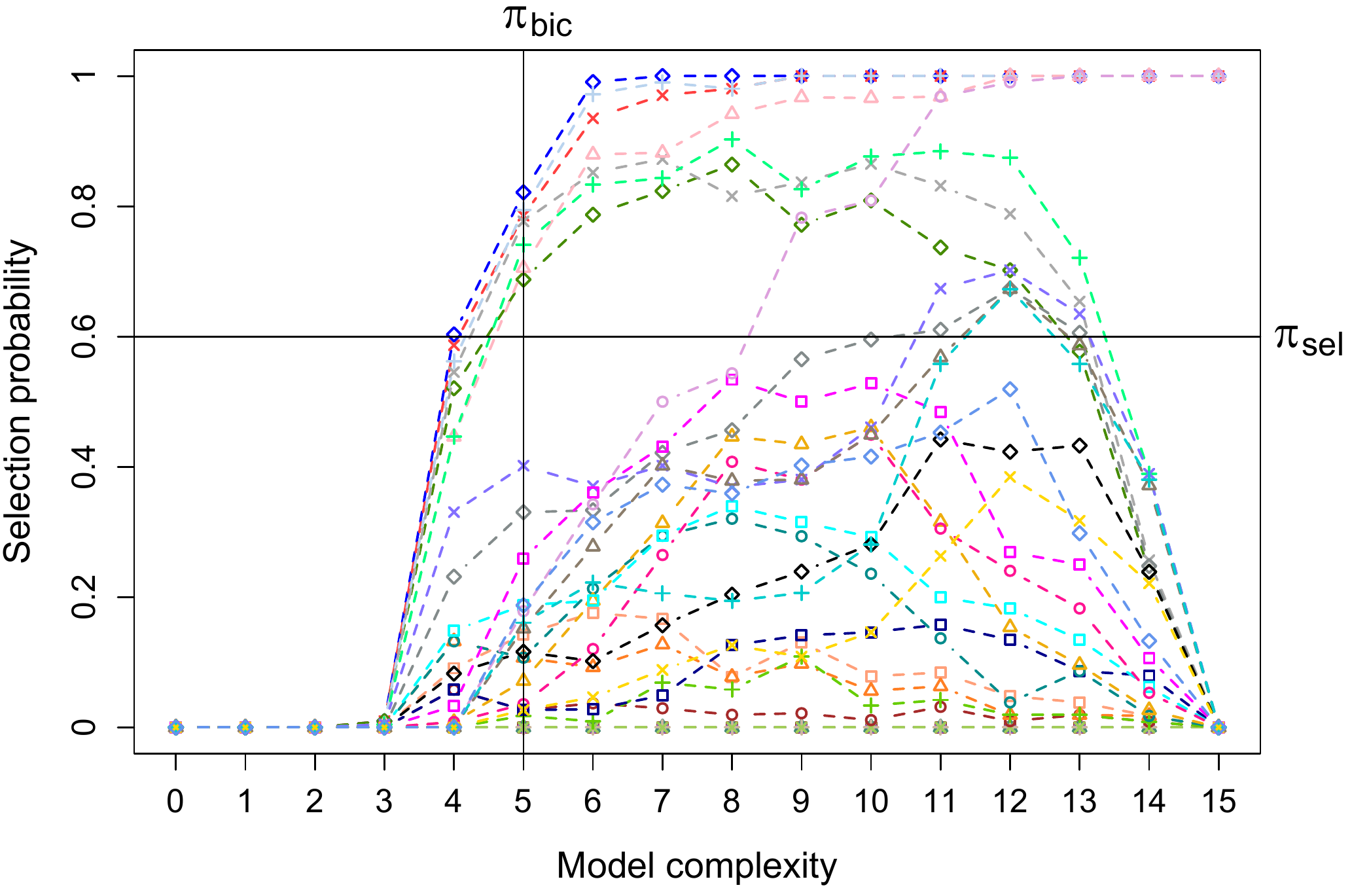}
    \label{causalStabADHD}
}}
\caption{The stability graphs for ADHD together with $\pi_{\mathrm{sel}}$ and $\pi_{\mathrm{bic}}$, yielding four regions. The top-left region is the area containing the relevant causal relations. (a) The edge stability graph showing four relevant edges. (b) The causal path stability graph showing seven relevant causal paths.
See Tables~\ref{tableRelevantEdgeADHD} and~\ref{tableRelevantCausalADHD} in Appendix \ref{sec:appendixA} for more detail.}
\label{stabADHD}
\end{figure}

In this experiment we consider a data set about \emph{Attention Deficit Hyperactivity Disorder} (ADHD) of $245$ subjects with $23$ variables\cite{cao2006abnormal}. Following \cite{sokolova2014causal}, we excluded instances with missing values and variables that either have insufficient instances or are considered irrelevant. The remaining data set consists of $221$ instances and six variables with $\mathrm{gender}$ the gender of subjects, $\mathrm{AD}$ the attention deficit measure, $\mathrm{HI}$ the assessment of hyperactivity/impulsivity symptoms, $\mathrm{aggression}$ the measure of aggressive behavior, $\mathrm{medication}$ the medication status of subjects, and $\mathrm{handedness}$ represents whether a subject uses the right and/or left hand. Following~\cite{rhemtulla2012can}, we treat all discrete variables as continuous variables. We added prior knowledge that the variable $\mathrm{gender}$ does not cause any of the other variables directly.

The total computation time for one subset was around $4.9$ minutes. Figure~\ref{stabADHD} shows that there are four relevant edges, namely between $\mathrm{gender}$ and $\mathrm{AD}$, $\mathrm{AD}$ and $\mathrm{medication}$, $\mathrm{AD}$ and $\mathrm{HI}$, and $\mathrm{HI}$ and $\mathrm{aggression}$.
Moreover, Figure~\ref{stabADHD} shows that there are seven relevant causal paths; $\mathrm{gender}$ to $\mathrm{AD}$, $\mathrm{gender}$ to $\mathrm{HI}$, $\mathrm{gender}$ to $\mathrm{medication}$, $\mathrm{gender}$ to $\mathrm{aggression}$, $\mathrm{AD}$ to $\mathrm{HI}$, $\mathrm{AD}$ to $\mathrm{medication}$, and $\mathrm{AD}$ to $\mathrm{aggression}$.

\begin{figure}[!t]
\centering
\includegraphics[width=0.45\textwidth]{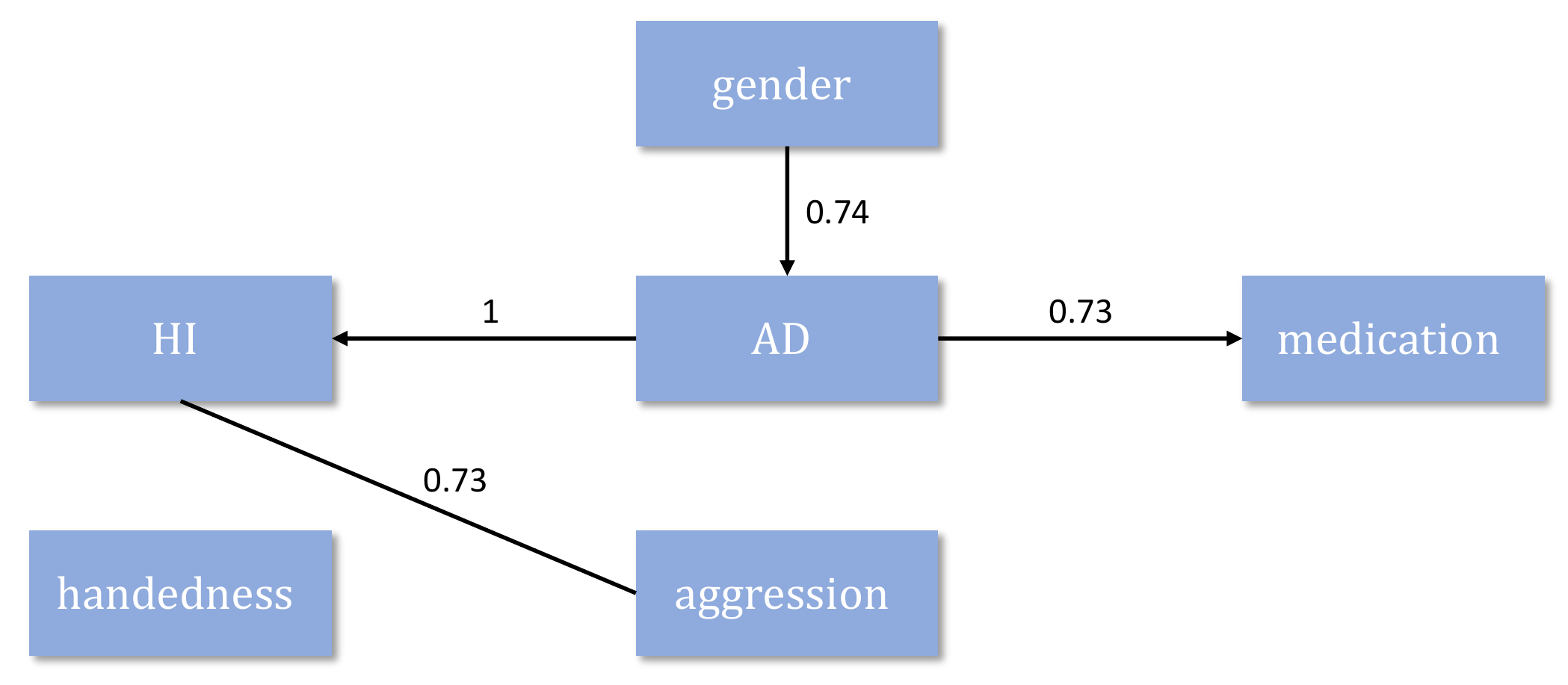} 
\caption{The inferred model of ADHD by combining the edge stability and causal path stability graphs. Each edge has a reliability score which is the highest selection probability in the top-left region of the edge stability graph}.
\label{ADHDModel}
\end{figure}

The stability graphs can be combined into a model as follows. First, the nodes are connected according to the four relevant edges obtained. Second, the edges are oriented according to the background knowledge added. The fact that the variable $\mathrm{gender}$ does not directly cause any other variable results in one directed edge $\mathrm{gender}\rightarrow \mathrm{AD}$. Third, the edges are oriented according to the relevant causal paths obtained. This results in two directed edges, $\mathrm{AD} \rightarrow \mathrm{HI}$ and $\mathrm{AD} \rightarrow \mathrm{medication}$. Since there is no relevant edge between $\mathrm{AD}$ and $\mathrm{aggression}$ and no relevant causal path from $\mathrm{HI}$ to $\mathrm{aggression}$ we cannot orient any other edges and therefore cannot represent two of the relevant causal paths in the model. We loose some information when converting the stability graphs into a model. The inferred model is shown in Figure~\ref{ADHDModel}.

The causal relations obtained for ADHD are corroborated by studies reported in the literature. In \cite{sokolova2014causal}, gender is shown to be a direct cause for attention deficit, attention deficit is shown to be a direct cause for both hyperactivity, medication, and aggression, and hyperactivity and aggression are related but neither variable is a direct cause for the other.

\section{Conclusion and Future Work}
\label{sec:conclusion}

In the last decades the field of causal modeling has seen a surge in theoretical development
and the construction of various causal discovery algorithms. In general, causal discovery algorithms
can be divided into two approaches: constraint-based and score-based. A disadvantage, however,
of current causal discovery algorithms is the inherent instability of structure estimation.
With finite samples small changes in the data can lead to completely different optimal structures.

The present work introduces a new hypothesis-free score-based causal discovery algorithm, stable specification search, that is robust for finite samples based on subsampling and selection algorithms. Our approach uses exploratory search to search over
Structural Equation Models and allows for the incorporation of prior background knowledge, without the need to specify the complete model structure in advance.

The comparison conducted on the simulated data shows that our method, the stable specification search, shows significant improvement over alternative approaches in obtaining the causal relations.
The results on both real-world data sets, CFS and ADHD, are consistent with previous studies \cite{vercoulen1998persistence,wiborg2012towards,heins2013process,sokolova2014causal}. In general we may conclude that our causal discovery algorithm is able to robustly estimate the underlying causal structure.

Several issues have not yet been explored in our current approach that warrant further research, such as
latent variables and longitudinal data. Taking into account the existence of latent variables can
further improve our structure estimate by properly identifying dependencies between variables as
an unmeasured common cause acting on both variables. In longitudinal data several subjects are measured
at different time slices which provides a richer structure that can be incorporated in the causal
discovery algorithm. A first attempt in this direction can be found in \cite{rahmadi2015causality}.

Our approach can be viewed as a novel application of multi-objective optimization. The main idea of stability selection~\cite{meinshausen2010stability}, is to increase the robustness of structure estimation by considering a whole range of model complexities. In the original work, this is done by varying a continuous regularization parameter. For causal discovery we have to explicitly consider different discrete model complexities. Furthermore, finding the optimal structure for each model complexity is a hard optimization problem. By rephrasing stability selection as a multi-objective optimization problem, we can jointly run over various model complexities and find the corresponding optimal structures for each model complexity. In this paper, we have used NSGA-{II} for multi-objective optimization, because of its popularity and availability, but realize that more recent multi-objective optimization approaches~\cite{qi2015immune,kukkonen2005gde3,zhang2007moea,taboada2008moms} may be even more efficient. This is beyond the scope of this work and left for future research. In the same spirit, one can easily combine freely available software packages, e.g., for scoring Structural Equation Models, bootstrap sampling, and multi-objective optimization, to build one's own robust structural estimation approach.

\ifCLASSOPTIONcompsoc
  \section*{Acknowledgments}
\else
  \section*{Acknowledgment}
\fi
The research leading to these results has received funding from the DGHE of Indonesia and the European Community's Seventh Framework Programme (FP7/2007-2013) under grant agreement n$^{\circ}$ 305697.

\bibliographystyle{IEEEtran}
\bibliography{Bibfile}
\appendices
\section{}
\label{sec:appendixA}
\clearpage

\begin{table}[!t]
\caption{Edge stability of CFS}
\label{tableRelevantEdgeCFS}
\centering
\begin{tabular}{c||c}
\hline
\bfseries Lines & \bfseries Edges \\
\hline\hline
    \includegraphics[page=1,scale=0.45]{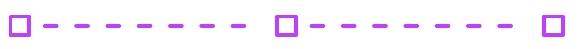} & $\mathrm{fatigue}$ and $\mathrm{pActivity}$\\ 
   \includegraphics[page=2,scale=0.45]{CFS/appendixCFSEdges} & $\mathrm{fatigue}$ and $\mathrm{control}$ \\ 
   \includegraphics[page=3,scale=0.45]{CFS/appendixCFSEdges} & $\mathrm{control}$ and $\mathrm{focusing}$   \\ 
   \includegraphics[page=4,scale=0.45]{CFS/appendixCFSEdges} & $\mathrm{fatigue}$ and $\mathrm{functioning}$   \\ 
   \includegraphics[page=5,scale=0.45]{CFS/appendixCFSEdges} & $\mathrm{control}$ and $\mathrm{functioning}$ \\ 
   \includegraphics[page=6,scale=0.45]{CFS/appendixCFSEdges} & $\mathrm{focusing}$ and $\mathrm{pActivity}$ \\ 
   \includegraphics[page=7,scale=0.45]{CFS/appendixCFSEdges} & $\mathrm{fatigue}$ and $\mathrm{focusing}$ \\ 
   \includegraphics[page=8,scale=0.45]{CFS/appendixCFSEdges} & $\mathrm{pActivity}$ and $\mathrm{oActivity}$ \\ 
   \includegraphics[page=9,scale=0.45]{CFS/appendixCFSEdges} & $\mathrm{functioning}$ and $\mathrm{pActivity}$ \\ 
   \includegraphics[page=10,scale=0.45]{CFS/appendixCFSEdges} & $\mathrm{control}$ and $\mathrm{pActivity}$ \\ 
   \includegraphics[page=11,scale=0.45]{CFS/appendixCFSEdges} & $\mathrm{control}$ and $\mathrm{oActivity}$ \\ 
   \includegraphics[page=12,scale=0.45]{CFS/appendixCFSEdges} & $\mathrm{focusing}$ and $\mathrm{oActivity}$ \\ 
   \includegraphics[page=13,scale=0.45]{CFS/appendixCFSEdges} & $\mathrm{fatigue}$ and $\mathrm{oActivity}$ \\ 
   \includegraphics[page=14,scale=0.45]{CFS/appendixCFSEdges} & $\mathrm{fucntioning}$ and $\mathrm{oActivity}$ \\ 
   \includegraphics[page=15,scale=0.45]{CFS/appendixCFSEdges} & $\mathrm{functioning}$ and $\mathrm{focusing}$ \\ 
   \hline
\end{tabular}
\end{table}

\begin{table}[!t]
\caption{Causal path stability of CFS}
\label{tableRelevantCausalCFS}
\centering
\begin{tabular}{c||c}
\hline
\bfseries Lines & \bfseries Causal Paths \\
\hline\hline
    \includegraphics[page=1,scale=0.45]{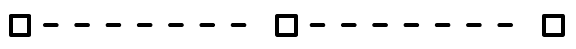} & $\mathrm{pActivity}$ to $\mathrm{fatigue}$\\ 
   \includegraphics[page=2,scale=0.45]{CFS/appendixCFSCausals} & $\mathrm{control}$ to $\mathrm{fatigue}$ \\ 
   \includegraphics[page=3,scale=0.45]{CFS/appendixCFSCausals} & $\mathrm{functioning}$ to $\mathrm{fatigue}$   \\ 
   \includegraphics[page=4,scale=0.45]{CFS/appendixCFSCausals} & $\mathrm{focusing}$ to $\mathrm{fatigue}$   \\ 
   \includegraphics[page=5,scale=0.45]{CFS/appendixCFSCausals} & $\mathrm{oActivity}$ to $\mathrm{fatigue}$ \\ 
   \includegraphics[page=6,scale=0.45]{CFS/appendixCFSCausals} & $\mathrm{focusing}$ to $\mathrm{pActivity}$ \\ 
   \includegraphics[page=7,scale=0.45]{CFS/appendixCFSCausals} & $\mathrm{functioning}$ to $\mathrm{pActivity}$ \\ 
   \includegraphics[page=8,scale=0.45]{CFS/appendixCFSCausals} & $\mathrm{oActivity}$ to $\mathrm{pActivity}$ \\ 
   \includegraphics[page=9,scale=0.45]{CFS/appendixCFSCausals} & $\mathrm{control}$ to $\mathrm{pActivity}$ \\ 
   \includegraphics[page=10,scale=0.45]{CFS/appendixCFSCausals} & $\mathrm{functioning}$ to $\mathrm{oActivity}$ \\ 
   \includegraphics[page=11,scale=0.45]{CFS/appendixCFSCausals} & $\mathrm{focusing}$ to $\mathrm{oActivity}$ \\ 
   \includegraphics[page=12,scale=0.45]{CFS/appendixCFSCausals} & $\mathrm{focusing}$ to $\mathrm{control}$ \\ 
   \includegraphics[page=13,scale=0.45]{CFS/appendixCFSCausals} & $\mathrm{control}$ to $\mathrm{oActivity}$ \\ 
   \includegraphics[page=14,scale=0.45]{CFS/appendixCFSCausals} & $\mathrm{functioning}$ to $\mathrm{control}$ \\ 
   \includegraphics[page=15,scale=0.45]{CFS/appendixCFSCausals} & $\mathrm{pActivity}$ to $\mathrm{oActivity}$ \\ 
   \includegraphics[page=16,scale=0.45]{CFS/appendixCFSCausals} & $\mathrm{control}$ to $\mathrm{functioning}$\\ 
   \includegraphics[page=17,scale=0.45]{CFS/appendixCFSCausals} & $\mathrm{oActivity}$ to $\mathrm{functioning}$ \\ 
   \includegraphics[page=18,scale=0.45]{CFS/appendixCFSCausals} & $\mathrm{oActivity}$ to $\mathrm{control}$   \\ 
   \includegraphics[page=19,scale=0.45]{CFS/appendixCFSCausals} & $\mathrm{oActivity}$ to $\mathrm{focusing}$   \\ 
   \includegraphics[page=20,scale=0.45]{CFS/appendixCFSCausals} & $\mathrm{control}$ to $\mathrm{focusing}$ \\ 
   \includegraphics[page=21,scale=0.45]{CFS/appendixCFSCausals} & $\mathrm{pActivity}$ to $\mathrm{functioning}$ \\ 
   \includegraphics[page=22,scale=0.45]{CFS/appendixCFSCausals} & $\mathrm{focusing}$ to $\mathrm{functioning}$ \\ 
   \includegraphics[page=23,scale=0.45]{CFS/appendixCFSCausals} & $\mathrm{functioning}$ to $\mathrm{focusing}$ \\ 
   \includegraphics[page=24,scale=0.45]{CFS/appendixCFSCausals} & $\mathrm{pActivity}$ to $\mathrm{control}$ \\ 
   \includegraphics[page=25,scale=0.45]{CFS/appendixCFSCausals} & $\mathrm{pActivity}$ to $\mathrm{focusing}$ \\ 
   \includegraphics[page=26,scale=0.45]{CFS/appendixCFSCausals} & $\mathrm{fatigue}$ to $\mathrm{pActivity}$ \\ 
   \includegraphics[page=27,scale=0.45]{CFS/appendixCFSCausals} & $\mathrm{fatigue}$ to $\mathrm{oActivity}$ \\ 
   \includegraphics[page=28,scale=0.45]{CFS/appendixCFSCausals} & $\mathrm{fatigue}$ to $\mathrm{focusing}$ \\ 
   \includegraphics[page=29,scale=0.45]{CFS/appendixCFSCausals} & $\mathrm{fatigue}$ to $\mathrm{functioning}$ \\ 
   \includegraphics[page=30,scale=0.45]{CFS/appendixCFSCausals} & $\mathrm{fatigue}$ to $\mathrm{control}$ \\ 
   \hline
\end{tabular}
\end{table}

\begin{table}[!t]
\caption{Edge stability of ADHD}
\label{tableRelevantEdgeADHD}
\centering
\begin{tabular}{c||c}
\hline
\bfseries Lines & \bfseries Edges \\
\hline\hline
    \includegraphics[page=6,scale=0.45]{ADHD/appendixEdgeADHD} & $\mathrm{AD}$ and $\mathrm{HI}$\\ 
   \includegraphics[page=7,scale=0.45]{ADHD/appendixEdgeADHD} & $\mathrm{AD}$ and $\mathrm{medication}$ \\ 
   \includegraphics[page=11,scale=0.45]{ADHD/appendixEdgeADHD} & $\mathrm{HI}$ and $\mathrm{aggression}$   \\ 
   \includegraphics[page=1,scale=0.45]{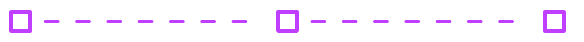} & $\mathrm{gender}$ and $\mathrm{AD}$   \\ 
   \includegraphics[page=8,scale=0.45]{ADHD/appendixEdgeADHD} & $\mathrm{aggression}$ and $\mathrm{AD}$ \\ 
   \includegraphics[page=10,scale=0.45]{ADHD/appendixEdgeADHD} & $\mathrm{AD}$ and $\mathrm{medication}$ \\ 
   \includegraphics[page=15,scale=0.45]{ADHD/appendixEdgeADHD} & $\mathrm{handedness}$ and $\mathrm{aggression}$ \\ 
   \includegraphics[page=13,scale=0.45]{ADHD/appendixEdgeADHD} & $\mathrm{medication}$ and $\mathrm{aggression}$ \\ 
   \includegraphics[page=2,scale=0.45]{ADHD/appendixEdgeADHD} & $\mathrm{gender}$ and $\mathrm{HI}$ \\ 
   \includegraphics[page=12,scale=0.45]{ADHD/appendixEdgeADHD} & $\mathrm{HI}$ and $\mathrm{handedness}$ \\ 
   \includegraphics[page=3,scale=0.45]{ADHD/appendixEdgeADHD} & $\mathrm{gender}$ and $\mathrm{medication}$ \\ 
   \includegraphics[page=5,scale=0.45]{ADHD/appendixEdgeADHD} & $\mathrm{gender}$ and $\mathrm{handedness}$ \\ 
   \includegraphics[page=9,scale=0.45]{ADHD/appendixEdgeADHD} & $\mathrm{AD}$ and $\mathrm{handedness}$ \\ 
   \includegraphics[page=4,scale=0.45]{ADHD/appendixEdgeADHD} & $\mathrm{gender}$ and $\mathrm{aggression}$ \\ 
   \includegraphics[page=14,scale=0.45]{ADHD/appendixEdgeADHD} & $\mathrm{medication}$ and $\mathrm{handedness}$ \\ 
   \hline
\end{tabular}
\end{table}

\begin{table}[!t]
\caption{Causal path stability of ADHD}
\label{tableRelevantCausalADHD}
\centering
\begin{tabular}{c||c}
\hline
\bfseries Lines & \bfseries Causal Paths \\
\hline\hline
    \includegraphics[page=6,scale=0.45]{ADHD/appendixCausalADHD} & $\mathrm{gender}$ to $\mathrm{AD}$\\ 
   \includegraphics[page=11,scale=0.45]{ADHD/appendixCausalADHD} & $\mathrm{gender}$ to $\mathrm{HI}$ \\ 
   \includegraphics[page=16,scale=0.45]{ADHD/appendixCausalADHD} & $\mathrm{gender}$ to $\mathrm{medication}$   \\ 
   \includegraphics[page=17,scale=0.45]{ADHD/appendixCausalADHD} & $\mathrm{AD}$ to $\mathrm{medication}$   \\ 
   \includegraphics[page=12,scale=0.45]{ADHD/appendixCausalADHD} & $\mathrm{AD}$ to $\mathrm{HI}$ \\ 
   \includegraphics[page=22,scale=0.45]{ADHD/appendixCausalADHD} & $\mathrm{AD}$ to $\mathrm{aggression}$ \\ 
   \includegraphics[page=21,scale=0.45]{ADHD/appendixCausalADHD} & $\mathrm{gender}$ to $\mathrm{aggression}$ \\ 
   \includegraphics[page=23,scale=0.45]{ADHD/appendixCausalADHD} & $\mathrm{HI}$ to $\mathrm{aggression}$ \\ 
   \includegraphics[page=18,scale=0.45]{ADHD/appendixCausalADHD} & $\mathrm{HI}$ to $\mathrm{medication}$ \\ 
   \includegraphics[page=19,scale=0.45]{ADHD/appendixCausalADHD} & $\mathrm{aggression}$ to $\mathrm{medication}$ \\ 
   \includegraphics[page=14,scale=0.45]{ADHD/appendixCausalADHD} & $\mathrm{aggression}$ to $\mathrm{HI}$ \\ 
   \includegraphics[page=7,scale=0.45]{ADHD/appendixCausalADHD} & $\mathrm{HI}$ to $\mathrm{AD}$ \\ 
   \includegraphics[page=24,scale=0.45]{ADHD/appendixCausalADHD} & $\mathrm{medication}$ to $\mathrm{aggression}$ \\ 
   \includegraphics[page=25,scale=0.45]{ADHD/appendixCausalADHD} & $\mathrm{handedness}$ to $\mathrm{aggression}$ \\ 
   \includegraphics[page=26,scale=0.45]{ADHD/appendixCausalADHD} & $\mathrm{gender}$ to $\mathrm{handedness}$ \\ 
   \includegraphics[page=30,scale=0.45]{ADHD/appendixCausalADHD} & $\mathrm{aggression}$ to $\mathrm{handedness}$\\ 
   \includegraphics[page=27,scale=0.45]{ADHD/appendixCausalADHD} & $\mathrm{AD}$ to $\mathrm{handedness}$ \\ 
   \includegraphics[page=28,scale=0.45]{ADHD/appendixCausalADHD} & $\mathrm{HI}$ to $\mathrm{handedness}$   \\ 
   \includegraphics[page=15,scale=0.45]{ADHD/appendixCausalADHD} & $\mathrm{handedness}$ to $\mathrm{HI}$   \\ 
   \includegraphics[page=20,scale=0.45]{ADHD/appendixCausalADHD} & $\mathrm{handedness}$ to $\mathrm{medication}$ \\ 
   \includegraphics[page=13,scale=0.45]{ADHD/appendixCausalADHD} & $\mathrm{medication}$ to $\mathrm{HI}$ \\ 
   \includegraphics[page=9,scale=0.45]{ADHD/appendixCausalADHD} & $\mathrm{aggression}$ to $\mathrm{gender}$ \\ 
   \includegraphics[page=29,scale=0.45]{ADHD/appendixCausalADHD} & $\mathrm{medication}$ to $\mathrm{handedness}$ \\ 
   \includegraphics[page=10,scale=0.45]{ADHD/appendixCausalADHD} & $\mathrm{aggression}$ to $\mathrm{HI}$ \\ 
   \includegraphics[page=8,scale=0.45]{ADHD/appendixCausalADHD} & $\mathrm{medication}$ to $\mathrm{AD}$ \\ 
   \includegraphics[page=1,scale=0.45]{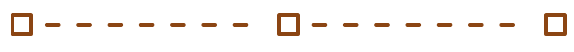} & $\mathrm{AD}$ to $\mathrm{gender}$ \\ 
   \includegraphics[page=2,scale=0.45]{ADHD/appendixCausalADHD} & $\mathrm{HI}$ to $\mathrm{gender}$ \\ 
   \includegraphics[page=3,scale=0.45]{ADHD/appendixCausalADHD} & $\mathrm{medication}$ to $\mathrm{gender}$ \\ 
   \includegraphics[page=4,scale=0.45]{ADHD/appendixCausalADHD} & $\mathrm{aggression}$ to $\mathrm{gender}$ \\ 
   \includegraphics[page=5,scale=0.45]{ADHD/appendixCausalADHD} & $\mathrm{handedness}$ to $\mathrm{gender}$ \\ 
   \hline
\end{tabular}
\end{table}

\end{document}